\pdfoutput=1
\documentclass{article}

\usepackage{microtype}
\usepackage{graphicx}
\usepackage{booktabs} % for professional tables
\usepackage[dvipsnames]{xcolor}

\usepackage[square,sort&compress,comma,numbers]{natbib}
\usepackage[utf8]{inputenc} % allow utf-8 input
\usepackage[T1]{fontenc}    % use 8-bit T1 fonts
\usepackage{url}            % simple URL typesetting
\usepackage{booktabs}       % professional-quality tables
\usepackage{amsfonts}       % blackboard math symbols
\usepackage{nicefrac}       % compact symbols for 1/2, etc.
\usepackage{microtype}      % microtypography
\usepackage{times}
\usepackage{soul}
\usepackage[utf8]{inputenc}
\usepackage{graphicx}
\usepackage{booktabs}
\usepackage{outlines}

\usepackage[colorlinks=true,
citecolor=blue,
filecolor=black,
linkcolor=blue,
urlcolor=blue]{hyperref}

\usepackage{microtype}
\usepackage{graphicx}
\usepackage{subfig}
\usepackage{booktabs} % for professional tables

\usepackage{multirow}
\usepackage{paralist}

\usepackage{booktabs} % for professional tables
\usepackage{multicol}
\usepackage{diagbox}

\usepackage{here}

\usepackage{amsmath,amssymb,amsfonts,amsbsy,amsfonts,latexsym}
\usepackage{multirow}
\usepackage{makecell}
\usepackage[labelfont=bf,textfont=it,belowskip=0pt,aboveskip=5pt,tableposition=top]{caption}
\usepackage{colortbl}
\usepackage{adjustbox}

\definecolor{colorA}{RGB}{189,201,225}
\definecolor{colorB}{RGB}{103,169,207}
\definecolor{colorC}{RGB}{ 28,144,153}
\definecolor{colorD}{RGB}{  1,108, 89}

\newcolumntype{R}{>{\columncolor{gray!40}}r}
\newcolumntype{L}{>{\columncolor{gray!40}}l}
\newcolumntype{C}{>{\columncolor{gray!40}}c}

\usepackage{tabularx,colortbl}
\usepackage{multirow}
\usepackage[normalem]{ulem}
\useunder{\uline}{\ul}{}

\usepackage{enumitem}

\usepackage{xparse}% http://ctan.org/pkg/xparse

\DeclareGraphicsExtensions{.pdf,.png}

\usepackage{longtable}
\usepackage{pgfplots}

\usepackage{amsmath}
\usepackage{pifont}% http://ctan.org/pkg/pifont

\usepackage{booktabs}       % professional-quality tables
\usepackage{amsfonts}       % blackboard math symbols
\usepackage{nicefrac}       % compact symbols for 1/2, etc.
\usepackage{microtype}      % microtypography
\usepackage{xspace}      % microtypography

\usepackage[utf8]{inputenc} % allow utf-8 input
\usepackage[T1]{fontenc}    % use 8-bit T1 fonts
\usepackage{chngcntr}

\usepackage{adjustbox}

\NewDocumentCommand{\var}{O{s} m O{}}{%
  \ensuremath{#1_{#2}^{#3}}% add \vphantom{<bizarre sup>}
}
\usepackage{siunitx}

\definecolor{light-gray}{gray}{0.80}

\newcommand\appref{Appendix~\ref}
\newcommand\eref{Eq.~\ref}
\newcommand\fref{Figure~\ref}
\newcommand\tref{Table~\ref}
\newcommand\sref{Section~\ref}

\newcommand\ha{ \rowcolor{orange!0}}

\newcommand\hc{ \rowcolor{orange!40}}

\newcommand\chb{\cellcolor{orange!15}}
\newcommand\chc{\cellcolor{orange!40}}

\definecolor{brickred}{rgb}{0.8, 0.25, 0.33}
\definecolor{brickred2}{rgb}{0.25, 0.8, 0.33}
\newcommand{\cm}{{\color{brickred2}{\ding{51}}}}
\newcommand{\xm}{{\color{brickred}{\ding{55}}}}
\usepackage[utf8]{inputenc}
\usepackage[english]{babel}

\newcommand{\loss}{\mathcal{L}}

\newcommand{\OURS}{\textsc{HAWQ-V3}\xspace}
\newcommand{\OURSD}{\textsc{HAWQV3+Dist}\xspace}

\usepackage[accepted]{icml2021}

\icmltitlerunning{\OURS: Dyadic Neural Network Quantization}

\begin{document}

\twocolumn[
\icmltitle{\OURS: Dyadic Neural Network Quantization}

% It is OKAY to include author information, even for blind
% submissions: the style file will automatically remove it for you
% unless you've provided the [accepted] option to the icml2021
% package.

% List of affiliations: The first argument should be a (short)
% identifier you will use later to specify author affiliations
% Academic affiliations should list Department, University, City, Region, Country
% Industry affiliations should list Company, City, Region, Country

% You can specify symbols, otherwise they are numbered in order.
% Ideally, you should not use this facility. Affiliations will be numbered
% in order of appearance and this is the preferred way.
\icmlsetsymbol{equal}{*}

\begin{icmlauthorlist}
\icmlauthor{Zhewei Yao}{equal,berkeley}
\icmlauthor{Zhen Dong}{equal,berkeley}
\icmlauthor{Zhangcheng Zheng}{equal,berkeley}
\icmlauthor{Amir Gholami}{equal,berkeley}\\
\icmlauthor{Jiali Yu}{amazon,sjtu}
\icmlauthor{Eric Tan}{berkeley}
\icmlauthor{Leyuan Wang}{amazon}
\icmlauthor{Qijing Huang}{berkeley}
\icmlauthor{Yida Wang}{amazon}
\icmlauthor{Michael W. Mahoney}{berkeley}
\icmlauthor{Kurt Keutzer}{berkeley}
\end{icmlauthorlist}

\icmlaffiliation{berkeley}{University of California, Berkeley}
\icmlaffiliation{sjtu}{Shanghai Jiao Tong University}
\icmlaffiliation{amazon}{Amazon}
\icmlcorrespondingauthor{Amir Gholami}{amirgh@berkeley.edu}

% You may provide any keywords that you
% find helpful for describing your paper; these are used to populate
% the "keywords" metadata in the PDF but will not be shown in the document
\icmlkeywords{Machine Learning, ICML}

\vskip 0.3in
]

% this must go after the closing bracket ] following \twocolumn[ ...

% This command actually creates the footnote in the first column
% listing the affiliations and the copyright notice.
% The command takes one argument, which is text to display at the start of the footnote.
% The \icmlEqualContribution command is standard text for equal contribution.
% Remove it (just {}) if you do not need this facility.

% \printAffiliationsAndNotice{}  % leave blank if no need to mention equal contribution
\printAffiliationsAndNotice{\icmlEqualContribution} % otherwise use the standard text.

\begin{abstract}
Current low-precision quantization algorithms often have the hidden
cost of conversion back and forth from floating point to quantized integer values.
This hidden cost limits the latency improvement realized by quantizing Neural Networks.
To address this, we present \OURS, a novel mixed-precision integer-only quantization framework.
The contributions of \OURS are the following:
(i)
An integer-only inference where the entire computational
graph is performed 
only with integer multiplication, addition, and bit shifting,
without any floating point operations or even integer division;
(ii)
A novel hardware-aware mixed-precision quantization method where the bit-precision
is calculated by solving an integer linear programming problem 
that balances the trade-off between model perturbation and other constraints, e.g., memory footprint and latency;
(iii) Direct hardware deployment and open source contribution for 4-bit uniform/mixed-precision
quantization in TVM, achieving an average speed up of $1.45\times$ for uniform 4-bit, as compared
to uniform 8-bit for ResNet50 on T4 GPUs; and
(iv) extensive evaluation of the proposed methods on
ResNet18/50 and InceptionV3, for various
model compression levels with/without mixed precision. 
For ResNet50, our INT8 quantization achieves an
accuracy of $77.58\%$, which is $2.68\%$
higher than prior integer-only work, and our mixed-precision INT4/8 quantization can
reduce INT8 latency by $23\%$ and still achieve $76.73\%$ accuracy. 
Our framework and the TVM implementation have been open sourced~\cite{HAWQ}.
\end{abstract}

\section{Introduction}
\label{sec:intro}

An important step toward realizing pervasive deep learning is enabling
real-time inference, both at the edge and in the cloud, with low energy consumption and 
state-of-the-art model accuracy.
This will have a significant impact on applications such as
real-time intelligent healthcare monitoring, autonomous driving,
audio analytics, and speech recognition.
Over the past decade, we have observed significant improvements in the
accuracy of Neural Networks (NNs) for various tasks.
However, the state-of-the-art models are often prohibitively large and too compute-heavy to be deployed for real-time use.
A promising approach to address this is through quantization~\cite{gray1998quantization,han2015deep},
where low-precision quantized integer
values are used to express the model parameters and feature maps.
That can help reduce the model footprint, and improve inference speed and energy consumption.

However, existing quantization algorithms often use \textit{simulated quantization}, where
the parameters are stored with quantization, but are cast to floating point for inference.
As a result, all or part of the inference operations (e.g. convolution, matrix operations, batch norm layers, residual connections)
are performed using floating point precision.
This of course limits the speed up as we cannot utilize low precision logic. 
To address this, we build upon existing integer-only quantization methods~\cite{jacob2018quantization}, and propose systematic methods to extend them to low and mixed-precision
quantization.
In particular, we make the following contributions:

\begin{itemize}
% [noitemsep, nolistsep, labelindent=0pt, leftmargin=*]
    \item
    We develop \OURS, a mixed-precision integer-only quantization framework with integer-only multiplication, addition, and bit shifting with static quantization.
    Importantly, no floating point and no integer division calculation is performed in the entire inference.
    This includes the batch norm layers and residual connections, which are typically kept at floating point~precision in prior integer-only quantization work~\cite{dong2019hawq}. While keeping these operations in floating point helps
    accuracy, this is not allowed for integer-only hardware.
    We show that ignoring this and attempting to deploy a model that uses floating point residual on integer-only hardware can lead to more than 90\% mismatch (~\fref{fig:accumulation}).
    \OURS completely avoids this by using a novel approach to perform residual connections in 
    pure integer-only arithmetic. 
    See~\sref{sec:residual_connection} and~\appref{sec:error_accumulation_of_fq} for details.
    
    \item
    We propose a novel hardware-aware mixed-precision quantization formulation that
    uses an Integer Linear Programming (ILP) problem to find the best bit-precision setting.
    The ILP solver
    minimizes the model perturbation while observing application-specific
    constraints on model size, latency, and total bit operations.
    Compared to the contemporary work of~\cite{hubara2020improving}, our approach
    is hardware-aware and uses direct hardware measurement to find a bit precision setting
    that has the optimal balance between latency and accuracy. 
    See~\sref{subsec:ilp} and~\appref{sec:ilp_interpolation} for details.
    
    \item
    To verify the feasibility of our approach, we deploy the quantized integer-only models using Apache TVM~\cite{chen2018tvm} for INT8, INT4, and mixed-precision 
    settings. To the best of our knowledge,
    our framework is the first that adds INT4 support to TVM. By profiling the latency of different layers, we show that we can achieve an average of $1.47\times$ speed up with INT4, as compared to INT8 on a T4 GPU for ResNet50. 
    See~\sref{subsec:hardware} and~\tref{tab:resnet18-50-constraint} for more details.
    
    \item
    We extensively test \OURS on a wide range of workloads, including ResNet18, ResNet50, and InceptionV3, and
    show that we can achieve a substantial performance improvement, as compared to the prior state-of-the-art. 
    For instance, we achieve an
    accuracy of $78.50\%$ with INT8 quantization, which is more than $4\%$
    higher than prior integer-only work for InceptionV3.
    Furthermore, we show that mixed-precision INT4/8 quantization can be used to achieve higher speed up as compared to INT8 inference with smaller impact on accuracy as compared to INT4 quantization. For example, for ResNet50 we can speedup latency
    by 23\% as compared to INT8 and still achieve 76.73\% accuracy.
    See~\sref{sec:results} and~\tref{tab:imagenet}, \ref{tab:resnet18-50-constraint} for more details.
\end{itemize}

%%%%%%%%%%%%%%%%%%%%%%%%%%%%%%%%%%%%%%%%%%%%%%%%%%%%%%%%%%%%%%%%%%%%%%%%%%%%%%%%%%%
\begin{figure*}[t]
\centering
\includegraphics[width=.49\textwidth]{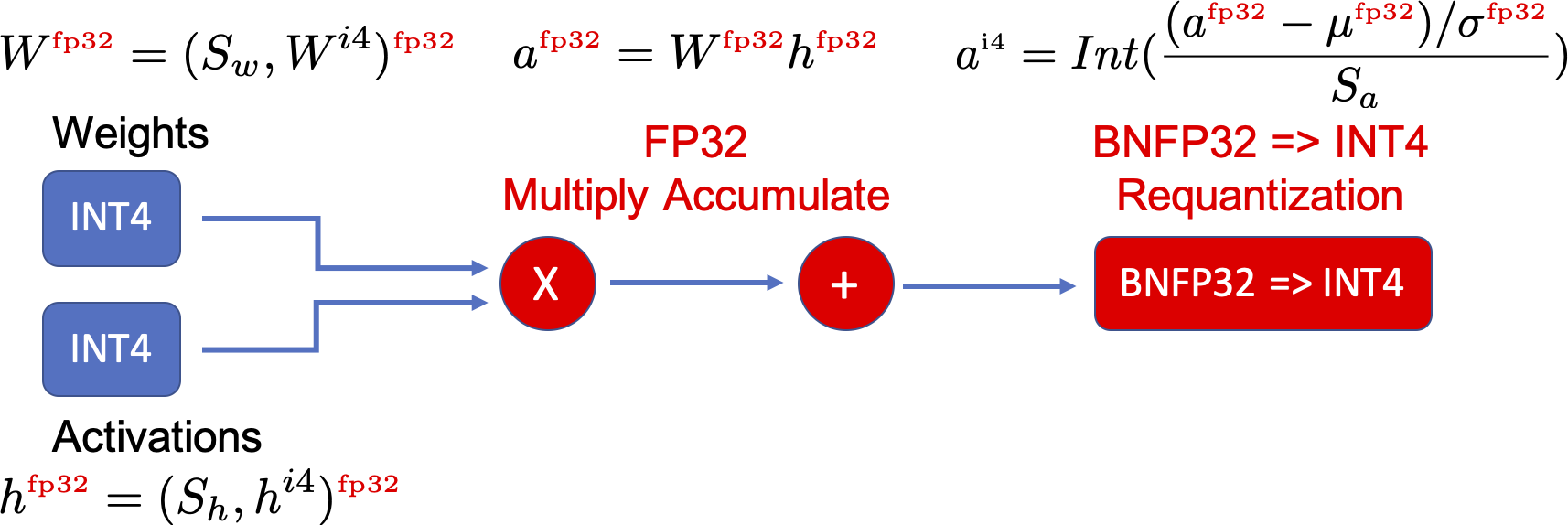}
\includegraphics[width=.44\textwidth]{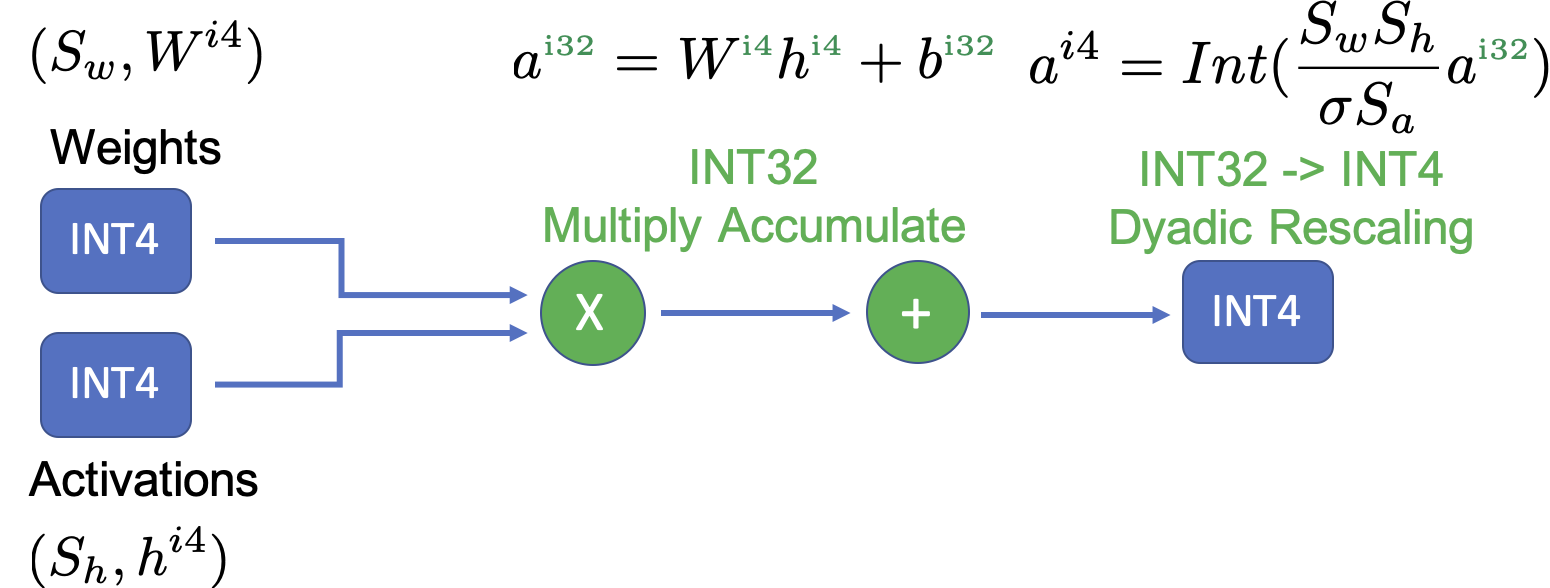}
\caption{
Illustration of fake vs true quantization for convolution and batch normalization folding. 
For simplicity, we ignore the affine coefficient of BN.  
(Left) In the simulated quantization (aka fake quantization approach), weights and activations
are simulated as integers with floating point representation, and all the multiplication and accumulation happen
in FP32 precision. Furthermore, the BN parameters (i.e. $\mu$ and $\sigma$) are stored and
computed using FP32 precision. This is undesirable but can significantly help accuracy since BN parameters
are sensitive to quantization.
However, with this approach, one cannot benefit from low-precision ALUs.
(Right) An illustration of the integer-only pipeline with dyadic arithmetic for convolution and BN folding.
The standard deviation ($\sigma$) in BN is merged into the quantization scale of the weights, and the mean
is quantized to INT32 and merged as a bias into the weights (denoted by $b^{\color{ForestGreen} i32}$)
Note that with this approach, all the weights and activations
are stored in integer format, and all the multiplications are performed with INT4 and accumulated in INT32 precision. 
Finally, the accumulated result is requantized to INT4 with dyadic scaling (denoted by $\frac{S_wS_h}{\sigma S_a}$). Importantly, no floating point or even
integer division is performed. 
See~\sref{subsec:batchnorm} and~\appref{sec:batchnorm_fusion} for more details.
}
  \label{fig:int32_batchnorm}
\end{figure*}
%%%%%%%%%%%%%%%%%%%%%%%%%%%%%%%%%%%%%%%%%%%%%%%%%%%%%%%%%%%%%%%%%%%%%%%%%%%%%%%%%%%

% --------------------------
\section{Related Work}
\label{sec:related_work}

There have been significant efforts recently to improve the trade-off between accuracy and efficiency of NN models.
These can be broadly categorized as follows:
(i) Designing new NN architectures~\cite{iandola2016squeezenet, sandler2018mobilenetv2, tan2019efficientnet};
(ii) Co-designing NN architecture and hardware together~\cite{han2017efficient,gholami2018squeezenext, wu2019fbnet, howard2019searching};
(iii) Pruning redundant filters~\cite{lecun1990optimal,han2015learning, molchanov2016pruning, li2016pruning, mao2017exploring, yang2017designing};
(iv) knowledge distillation~\cite{hinton2015distilling, mishra2017apprentice, polino2018model, yin2020dreaming};
and (v) using quantization (reduced precision).
Here, we provide a more detailed overview of this related work.

\textbf{Quantization.} 
A common solution is to compress NN models with quantization~\cite{asanovic1991experimental, hubara2016binarized, rastegari2016xnor, zhou2017incremental, zhou2016dorefa, jacob2018quantization, zhang_2018_lqnets, dong2019hawq, chin2020one, song2020drq,bhalgat2020lsq, kim2021zero, sharma2018bit,park2018energy},
where low-bit precision is used for weights/activations.
Quantization reduces model size without changing the original network architecture,
and it could potentially permit the use of low-precision matrix multiplication or convolution.

While the gains on speed/power increase for low-precision quantization, low-precision quantization suffers from accuracy degradation.
To address this, recent work uses 
non-uniform quantizers~\cite{zhang_2018_lqnets}, channel-wise 
quantization~\cite{krishnamoorthi2018whitepaper}, and progressive quantization-aware 
fine-tuning~\cite{zhou2017incremental}.
Other works try to include periodic regularization to assist 
quantization~\cite{naumov2018periodic, elthakeb2020gradient}, apply post training 
quantization~\cite{banner2019post, cai2020zeroq, hubara2020improving}, or improve 
accuracy by changing the channel counts accordingly for different 
layers~\cite{chin2020one}.
Despite these advances, performing uniform ultra low-bit quantization still results in a
significant accuracy degradation.
A promising direction is to use mixed-precision quantization~\cite{zhou2017adaptive, wang2018haq, dong2019hawq, shen2020q}, 
where some layers are kept at higher precision, while others are kept at a lower precision.
However, a challenge with this approach is finding the right
the mixed-precision setting for the different layers.
A brute force approach is not feasible since the search space is exponentially
large in the number of layers.

HAQ~\cite{wang2018haq} proposes to search this space by applying a 
Reinforcement Learning algorithm, while~\cite{wu2018mixed} uses a Differentiable 
Neural Architecture Search.
However, these searching methods require large computational 
resources, and their performance is very sensitive to hyper-parameters and even
initialization.
To address these issues, HAWQ~\cite{dong2019hawq,dong2019hawqv2} introduces an automatic way to find
good mixed-precision settings based on the sensitivity obtained using the Hessian spectrum. 
However, the Pareto frontier method in~\cite{dong2019hawqv2} is not flexible enough to satisfy simultaneously different requirements on hardware.
To address this, we propose here an ILP solution that can generate mixed-precision settings with various constraints (such as model size, BOPS, and latency), and which can be solved within seconds on commodity hardware.
The contemporary work of~\cite{hubara2020improving} also proposes to use an ILP. 
However, their approach is not hardware aware, and their approach uses FP32 casting.

Another issue is that the quantized weights and
activations need to be converted to floating point precision during inference, as shown in~\fref{fig:int32_batchnorm}.
This high-precision casting can have high overhead and limits inference speed, especially for hardware with limited on-chip memory. Using FP32 ALUs also requires a larger die area
in the chip, further limiting the peak computational capacity of the hardware.
The work of~\cite{jacob2018quantization} addresses
this casting problem by using integer-only quantization in INT8 precision.
However, there are several
shortcomings associated with their approach (which are addressed in \OURS).
First, \cite{jacob2018quantization} does not support
low-precision or mixed-precision quantization.
We show that this is useful in practice, as it
can improve the inference speed by up to 50\% 
with a small impact on accuracy.
Second,
both~\cite{jacob2018quantization} and HAWQ are hardware agnostic and do not
co-design/adapt the quantization for the target hardware.
In contrast, the ILP approach in \OURS is hardware aware, and it directly takes this into account
when determining mixed-precision bit setting.
Third, as we discuss in~\sref{subsec:batchnorm}, the approach used in~\cite{jacob2018quantization}
leads to sub-optimal accuracy for INT8 quantization, 
while our approach can achieve up to 5\% higher accuracy for INT8 inference.
Finally, to address the absence of low-precision support in previous works~\cite{jacob2018quantization,dong2019hawq}, we extend TVM to support INT4 
and mixed-precision quantization, and we validate our results by directly running the quantized model with low bit-width on the hardware. 
See~\appref{sec:deplotment_frameworks} for the discussion of different deployment frameworks.

\section{Methodology}
\label{sec:methodology}

Assume that the NN has $L$ layers with learnable parameters, denoted as $\{W_1, W_2, ..., W_L\}$, with $\theta$ denoting the combination of all such parameters. 
For a supervised setting, the goal is to optimize the following empirical risk minimization loss function:
\begin{equation}
\small
\label{eq:loss_function}
    \loss(\theta) = \frac1N\sum\nolimits_{i=1}^N l(x_i, y_i; \theta),
\end{equation}
where $(x, y)$ is the input data and the corresponding label, $l(x,y;\theta)$ is the loss function (e.g., MSE or Cross Entropy loss),
and $N$ is the total number of data points. 
We assume that we have the trained model parameters $\theta$ given
in floating point precision. Our goal is to quantize the model with the optimal trade-offs among
memory footprint, speed, and accuracy. 
Below, we first define quantization and then present \OURS.

\textbf{Uniform Quantization.} 
Quantization restricts NN weights/activations to a finite set of values as follows:
\begin{equation}
\small
\label{eq:quantization_formula}
Q(r) = \text{Int}\big({r}/{S}\big)-Z,
% Sq_j,~~~\text{for } z \in (t_j , t_{j+1}],
\end{equation}
where $Q$ is the quantization operator, $r$ is a real valued number (activation or a weight),
$S$ is a real valued scaling factor, and $Z$ is the zero point,
chosen such that the $0$ value would exactly map to quantized values.
Furthermore, $\text{Int}$ maps a floating point value to an integer value 
through a rounding operation (e.g., round to nearest and
truncation).

This formulation for $Q$ corresponds to uniform quantization.
However, some work in the literature has also explored non-uniform
quantization~\cite{zhang_2018_lqnets,park2018value,wang2018haq}. 
Although non-uniform quantization may achieve higher accuracy for a fixed bit-width, such approaches
are typically difficult to deploy on hardware to reduce latency.\footnote{However, they can reduce total model footprint.}
As such, for \OURS, we only focus on uniform quantization.
Meanwhile, for \OURS, we use (i) symmetric quantization for weights and asymmetric quantization for activations; and (ii) static quantization for all the scaling factors $S$. 
Meanwhile, we apply channel-wise quantization for different convolutional output channels.
Please see Appendix~\ref{sec:quantization_method} for more details. 

% -----------------------------------------------------
\subsection{Quantized Matrix Multiplication and Convolution}
\label{sec:quantized_matmul}

Consider a layer with hidden activation denoted as $h$ and weight tensor denoted as $W$,
followed by ReLU activation.
First, $h$ and $W$ are quantized to $S_hq_h$ and $S_wq_w$,
where $S_h$ and $S_w$ are the real valued quantization scales,
$q_h$ and $q_W$ are the corresponding quantized integer values.
The output result, denoted with $a$, can be computed as follows:
\begin{equation}
\small
\label{eq:quantized_matmul}
    a = S_wS_h (q_w * q_h),
\end{equation}
where $q_w*q_h$ is the matrix multiplication (or convolution) 
calculated with integer in low precision (e.g., INT4) and accumulated in INT32 precision.
%As such, $a$ will be in INT32 precision. 
This result is then requantized
and sent to the next layer as follows:
\begin{equation}
\small
\label{eq:requantize_full}
q_a =\text{Int}\left(\frac{a}{S_a}\right) = \text{Int}\left(\frac{S_wS_h}{S_a}(q_w * q_h)\right),
\end{equation}
where $S_a$ is the pre-calculated scale factor for the output activation.

In \OURS, the $q_w*q_h$ operation is performed with low-precision
integer-only multiplication and INT32 accumulation, and the final
INT32 result is quantized by scaling it with ${S_wS_h}/{S_a}$.
The latter is a floating point scaling that needs to be multiplied
with the accumulated result (in INT32 precision).
A naive implementation requires floating point multiplication for this stage.
However, this can be avoided by enforcing the scaling to be a
dyadic number. Dyadic numbers are rational numbers with the format of ${b}/{2^c}$, 
where $b,~c$ are two integer numbers.
As such, a dyadic scaling in~\eref{eq:requantize_full} can be efficiently performed
using INT32 integer multiplication and bit shifting. 
Given a specific ${S_wS_h}/{S_a}$, we use $DN$ (representing Dyadic Number) to denote the function that can calculate the corresponding $b$ and $c$:
\begin{equation}
\small
    {b}/{2^c} = \text{DN}\left({S_wS_h}/{S_a}\right).
\end{equation}
An advantage of using dyadic numbers besides avoiding floating point arithmetic, is that it removes the need to support division (which typically has an order of magnitude higher latency than multiplication) in the hardware. 
This approach is used for INT8 quantization in~\cite{jacob2018quantization},
and we enforce all the rescaling to be dyadic for low-precision and mixed-precision
quantization as well.

% ----------------------------------------
\subsection{Batch Normalization}
\label{subsec:batchnorm}

Batch normalization (BN) is an important component of most NN architectures, especially for computer vision applications.
BN performs the following
operation to an input activation $a$:
\begin{equation}
\small
\label{eq:batch_normalization}
    \text{BN}(a) = \beta \frac{a - \mu_B}{\sigma_B} + \gamma 
\end{equation}
where $\mu_B$ and $\sigma_B$ are the mean and standard deviation of $a$,  and $\beta,\ \gamma$ are trainable parameters. 
During inference, these parameters (both statistics and trainable parameters) are fixed, and
therefore the BN operations could be fused with the convolution (see Appendix~\ref{sec:batchnorm_fusion}).
However, an important problem is that quantizing the BN parameters often leads to significant accuracy degradation.
As such, many prior quantization methods keep BN parameters in FP32 precision (e.g.,~\cite{dong2019hawqv2, cai2020zeroq, chin2020one, choi2018pact, zhang_2018_lqnets, park2018value}, just to name a few). 
This makes such approaches not suitable for integer-only hardware.
While using such techniques help accuracy, \OURS completely avoids that.
We fuse the BN parameters with the convolution and quantized them
with integer-only approach
(Please see~\fref{fig:int32_batchnorm} where we compare simulated qauntization and \OURS
for BN and convolution.).

Another important point to discuss here is that we found the BN folding used
in~\cite{jacob2018quantization} to be sub-optimal.
In their approach  BN and CONV layers are fused together while BN running 
statistics are still kept updating.
% the running statistics of the BN layer. 
This actually requires computing each convolution layer
twice, once without BN and then with BN (as illustrated in~\citep[Figure C8]{jacob2018quantization}).
However, we found that this is unnecessary and degrades the accuracy.
Instead, in \OURS, we follow a simpler approach where we first keep the Conv and BN layer unfolded, and allow the BN statistics to update. 
After several epochs, we then freeze the running statistics in the BN layer and fold the CONV and BN layers (please see Appendix~\ref{sec:batchnorm_fusion} for details).
As we will show in~\sref{sec:results}, this approach results in better accuracy as compared to~\cite{jacob2018quantization}.

%%%%%%%%%%%%%%%%%%%%%%%%%%%%%%%%%%%%%%%%%%%%%%%%%%%%%%%%%%%%%%%%%%%%%%%%%%%%%%%%%
\begin{figure}[t]
\centering
\includegraphics[height=0.33\textwidth]{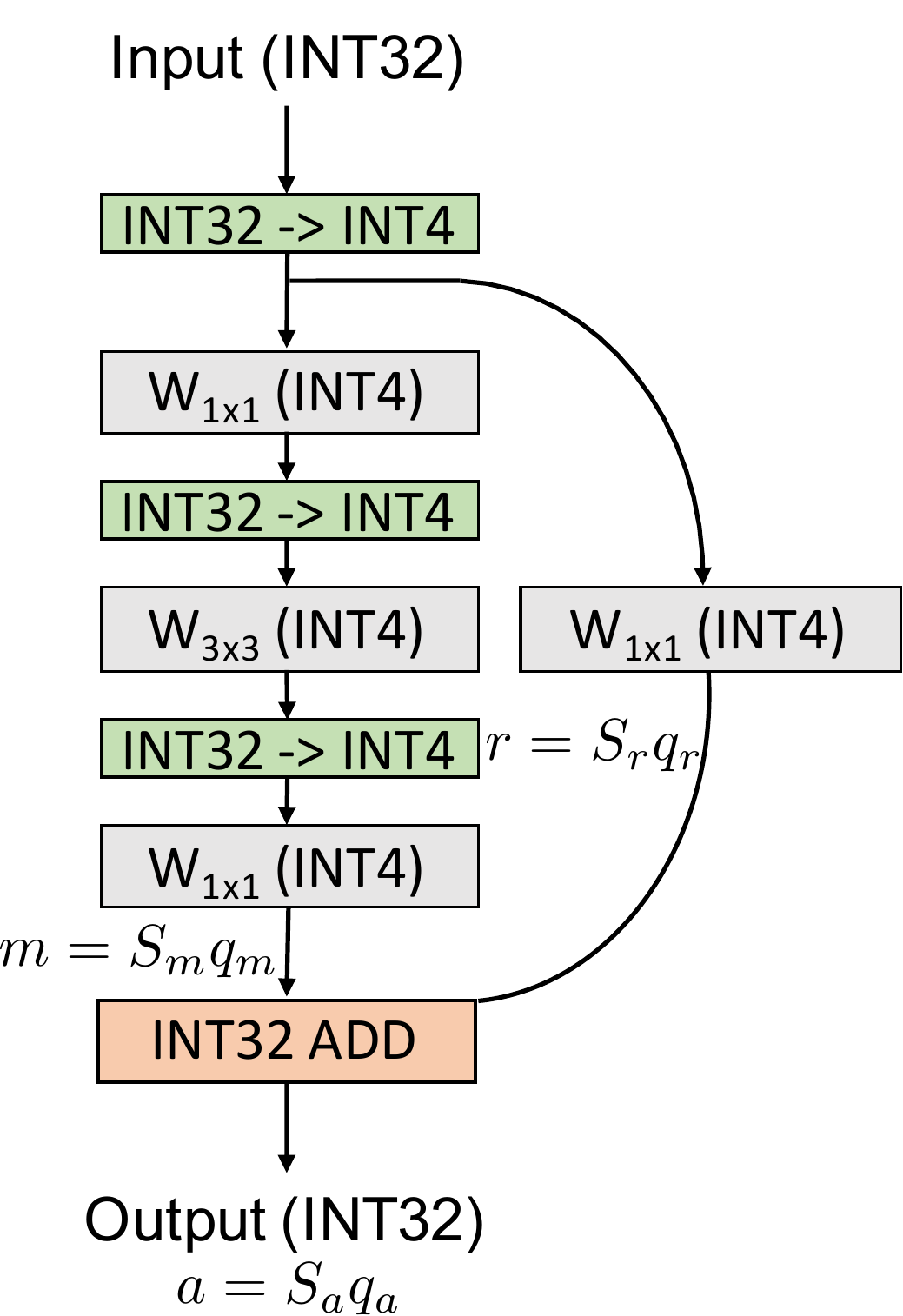}
\includegraphics[height=0.33\textwidth]{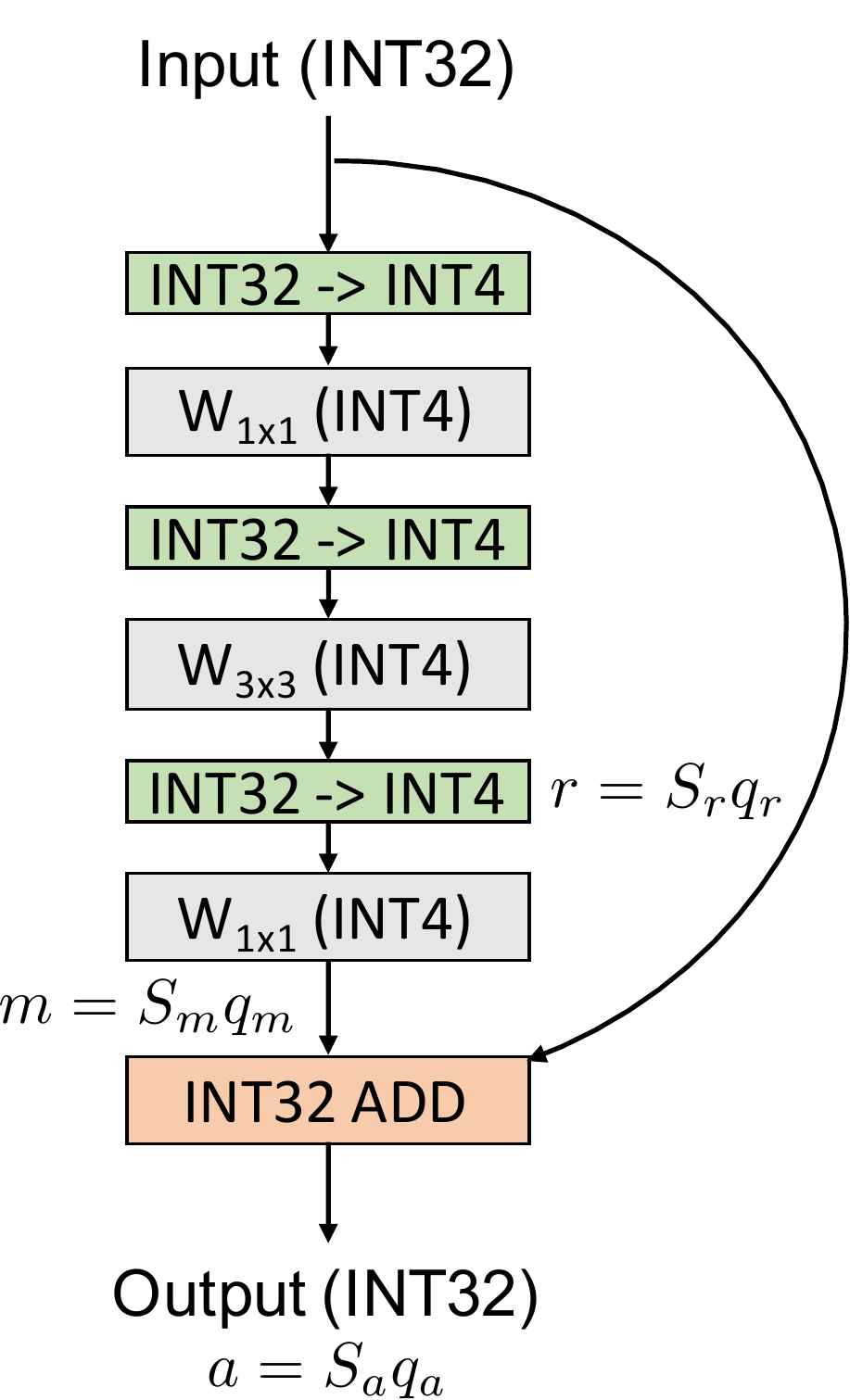}
\caption{Illustration of \OURS for a residual block with and without transition layer. 
Input feature map is given in INT32 precision, which is requantized to
INT4 precision (green boxes) before any convolution layer (gray boxes).
The BN layer is folded into the convolution.
The residual addition is performed in INT32 precision, and the final
accumulated result is re-scaled and sent to the next layer.
For blocks with a transition layer, we only quantize the input once to
INT4 and we use the same result for both $1\times1$ convolutions.
}
\label{fig:integer_only_resnet}
\end{figure}
%%%%%%%%%%%%%%%%%%%%%%%%%%%%%%%%%%%%%%%%%%%%%%%%%%%%%%%%%%%%%%%%%%%%%%%%%%%%%%%%%

% --------------------------------------------
\subsection{Residual Connection}
\label{sec:residual_connection}

Residual connection~\cite{he2016deep} is another important component in many NN architectures.
Similar to BN, quantizing the residual connections can lead to accuracy degradation,
and as such, some prior quantization works perform the operation in FP32 precision~\cite{choi2018pact, zhang_2018_lqnets, wang2018haq}. 
There is a common misunderstanding that this may not be a big problem.
However, this actually leads to complete loss of signal, especially for low
precision quantization. The main reason for this is that 
quantization is not a linear operation, that is $Q(a+b)\neq Q(a) + Q(b)$ ($a$, $b$ are floating point numbers).
As such, performing the accumulation in FP32 and then quantizing
is not the same as accumulating quantized values. 
Therefore, it is not possible to deploy quantization methods that keep
residual connection in FP32 in integer-only hardware (we provide more detailed
discussion of this in~\appref{sec:fake_res_con_layer} and also 
quantify the resulting error which can be more than 90\%).

We avoid this in \OURS, and use INT32 for the residual branch.
We perform the following steps to
ensure that the addition operation can happen with dyadic arithmetic.
Let us denote the activation passing through the residual connection as
$r=S_{r}q_{r}$.\footnote{This is either the input or the output activation after
the downsampling layer.}
Furthermore, let us denote the activation of the main branch before residual addition as
$m=S_{m}q_{m}$, and the final output after
residual accumulation by $a=S_{a}q_{a}$.
Then we will have:

\begin{equation}
\small
    q_{a} = \text{DN}\left({S_{m}}/{S_{a}}\right) q_m + \text{DN}\left({S_{r}}/{S_{a}}\right) q_r.
\end{equation}
Note that with this approach, we only need to perform a dyadic scaling of
$q_m$ and add the result with the dyadically scaled $q_r$. All of these
operations can happen with integer-only arithmetic.
Also we should note that in our approach all the scales are statically known.
These steps are schematically illustrated in~\fref{fig:integer_only_resnet} for
a residual connection with/without downsampling. 
Similar approach is performed for concatenation layer as well (see~\appref{sec:concatenation_layer}).

%%%%%%%%%%%%%%%%%%%%%%%%%%%%%%%%%%%%%%%%%%%%%%%%%%%%%%%%%%%%%%%%%%%%%%%%%%%%%%%%%%%
\begin{figure}[t]
\centering
\includegraphics[width=.5\textwidth]{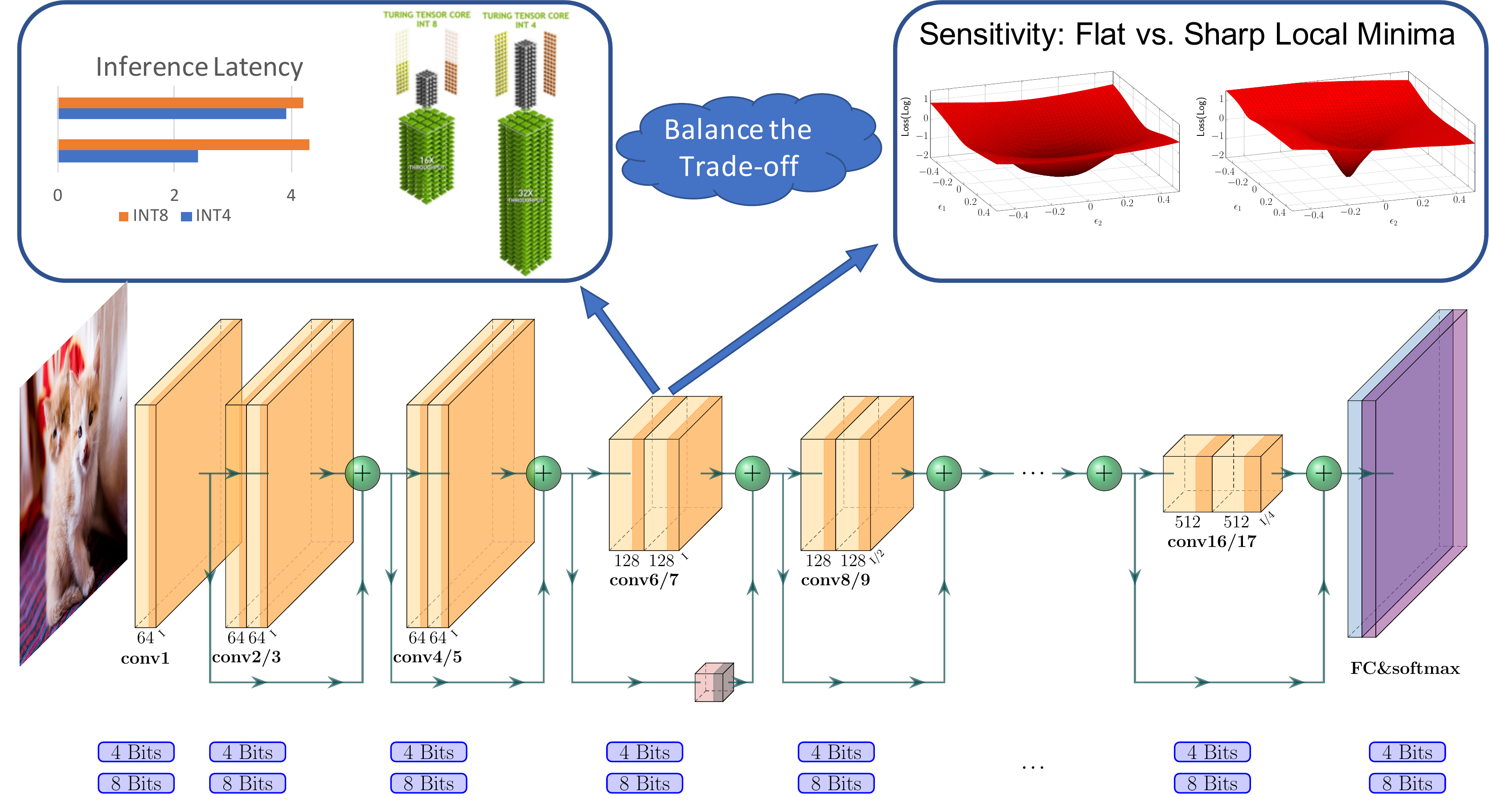}
\caption{
Illustration of inference speed and generalization performance trade-off of ResNet18. 
For each layer, we need to consider the speedup of INT4 vs INT8 and the sensitivity based on the second order (Hessian) sharpness~\cite{dong2019hawqv2} of this layer.
}
  \label{fig:resnet18_mixed_precision}
\end{figure}
%%%%%%%%%%%%%%%%%%%%%%%%%%%%%%%%%%%%%%%%%%%%%%%%%%%%%%%%%%%%%%%%%%%%%%%%%%%%%%%%%%%
% -----------------------------------
\subsection{Mixed Precision and Integer Linear Programming}
\label{subsec:ilp}

Uniformly quantizing all the layers to low
bit-width (e.g. INT4) could lead to significant accuracy degradation. 
However, it is possible
to benefit from low-precision quantization by keeping a subset of sensitive layers
at high precision~\cite{dong2019hawq}.
The basic idea is to keep sensitive layers at higher precision and insensitive layers at lower precision.
An important component of \OURS is that we directly consider hardware-specific metrics
such as latency, to select the bit-precision configuration. 
This is important since a layer's latency does not necessarily halve when quantized from INT8 to INT4 precision. 
In fact, as we discuss in~\sref{sec:results}, there are specific
layer configurations that do not gain any speed up when quantized to low precision,
and some that superlinearly benefit from quantization.\footnote{The speedup of each layer is calculated by the latency of INT8 divided by that of INT4. For uniform 4-bit and mixed-precision models, the speedup is calculated related to uniform 8-bit model.}
As such, quantizing the former will not lead to any latency improvement, and will only hurt accuracy. Therefore, it is better to keep such layers at high precision,
even if they have low sensitivity.
These trade-offs between accuracy and latency should be taken into consideration when quantizing them to low precision.
Importantly, these trade-offs are hardware-specific as latency in general does not correlate with the model size and/or FLOPS.
However, we can consider this by
directly measuring the latency of executing a layer in quantized precision on the target hardware platform.
This trade-off is schematically shown in~\fref{fig:resnet18_mixed_precision} (and
later quantified in~\fref{fig:resnet18_conv_speedups}).
We can use an Integer Linear Programming (ILP) problem to formalize the problem definition of finding
the bit-precision setting that has optimal trade-off as described~next.

Assume that we have $B$ choices
for quantizing each layer (i.e., 2 for INT4 or INT8). For a 
model with $L$ layers, the search space of the ILP will be $B^L$. 
The goal of solving the ILP problem is to find the best bit configuration among these
$B^L$ possibilities that results in optimal trade-offs between model perturbation $\Omega$,
and user-specified constraints such as model size, BOPS, and latency.
Each of these bit-precision 
settings could result in a different model perturbation.
To make the problem
tractable, we assume that the perturbations for each layer are independent of each
other (i.e., $\Omega = \sum_{i=1}^L\Omega_i^{(b_i)}$, where $\Omega_i^{(b_i)}$ is the $i$-th layer's perturbation with $b_i$ bit)\footnote{Similar assumption can be found in~\cite{dong2019hawq,dong2019hawqv2}.}.
This allows us to precompute the sensitivity of each
layer separately, and it only requires $BL$ computations.
For the sensitivity metric, we use the Hessian based perturbation proposed in~\citep[Eq. 2.11]{dong2019hawqv2}.
The ILP problem tries to find the right bit precision that minimizes this sensitivity, as follows:
\begin{align}
\small
\label{eq:ilp}
    \text{Objective: } &\min\nolimits_{\{b_i\}_{i=1}^L} \sum\nolimits_{i=1}^L \Omega_i^{(b_i)},\\
    \small
    \text{Subject to: } & \sum\nolimits_{i=1}^L M_i^{(b_i)} \leq\text{ Model Size Limit}, \\
    \small
                        & \sum\nolimits_{i=1}^L G_i^{(b_i)} \leq\text{ BOPS Limit}, \\
    \small
                        & \sum\nolimits_{i=1}^L Q_i^{(b_i)} \leq\text{ Latency Limit}.
\end{align}
Here, $M_i^{(b_i)}$ denotes the size of $i$-th layer with $b_i$ bit quantization,
$Q_i^{(b_i)}$ is the associated latency, and
$G_i^{(b_i)}$ is the corresponding BOPS required for computing that layer.
The latter measures the total Bit Operations for calculating a layer~\cite{van2020bayesian}:
\begin{equation*}
\small
    G_i^{(b_i)} = b_{w_i}b_{a_i} \text{MAC}_i,
\end{equation*}
where MAC$_i$ is the total Multiply-Accumulate operations for computing the $i$-th layer,
and $b_{w_i},\ b_{a_i}$ are the bit precision used for weight and activation.\footnote{$b_{w_i}$ and $b_{a_i}$ are always the same in \OURS. 
As such, \OURS does not need to cast lower-precision integer numbers, e.g., INT4, to higher-precision integer numbers, e.g., INT8, which is more efficient than~\cite{dong2019hawqv2,cai2020zeroq,wang2018haq}.}
Note that it is not necessary to set all these constraints at the same time. 
Typically, which constraint to use depends on the end-user application.

We solve the ILP using open source PULP library~\cite{pulp} in Python,
where we found that for all the configurations tested in the paper, the ILP solver can find the solution
in less than 1 second given the sensitivity metric.
For comparison, the RL based method of~\cite{wang2018haq} could take tens of hours
to find the right bit-precision setting.
Meanwhile, as can be seen, our ILP solver can be easily used for multiple constraints. 
However, the complexity of Pareto frontier proposed by~\cite{dong2019hawqv2} is exponentially increasing for multiple constraints.
In~\sref{sec:mixed_precision_with_diff_constraints}, we show the results with different constraints. 

We should also mention that the contemporary work of~\cite{hubara2020improving},
also proposed an ILP formulation. However, our approach is hardware-aware and we directly
deploy and measure the latency of each layer in hardware.
% -----------------------------------------------
\subsection{Hardware Deployment}
\label{subsec:hardware}

Model size alone is not a good metric to measure the efficiency (speed and energy consumption) of NNs.
In fact, it is quite possible that a small model would have higher latency and consume 
a larger amount of energy for inference. 
The same is also true for FLOPs. 
The reason is that neither model size nor FLOPs can account for cache misses, data locality, memory bandwidth, underutilization of hardware, etc. 
To address this, we need to deploy and directly measure the latency.

We target Nvidia Turing Tensor Cores of T4 GPU for deployment, as it supports both INT8 and INT4 precision and has been enhanced for deep learning network inference.
The only API available is the WMMA kernel call which is a micro-kernel for performing matrix-matrix operations in INT4 precision on Tensor Cores.
However, there is also no existing compiler that would map a NN quantized to INT4 to Tensor Cores using WMMA instructions.
To address this challenge, another contribution of our work is extending TVM~\cite{chen2018tvm} 
to support INT4 inference with/without mixed precision with INT8.
This is important so we can verify the speed benefits of mixed-precision inference.
To accomplish this, we had to
add new features in both graph-level IR and operator schedules to make INT4 inference efficient.
For instance, when we perform optimizations such as  memory planning, constant folding,
and operator fusion, at the graph-level IR, 4-bit data are involved.
However, on byte-addressable machines, manipulating 4-bit data individually leads to inefficiency in storage and communication. Instead, we pack eight 4-bit elements into an INT32 
data type and perform the memory movement as a chunk. In the final code generation stage, the data type and all memory access will be adjusted for INT32.
By adopting similar scheduling strategies to Cutlass~\cite{cutlass}, we implement a new direct convolution 
schedule for Tensor Cores for both 8-bit and 4-bit data in TVM.
We set the knobs for the configurations such as thread size, block size, and loop ordering so 
that the auto-tuner in TVM could search for the best latency settings.

Another important point is that we have completed the pipeline to test directly the trained weights and to avoid using random weights for speed measurements.
This is important, since small discrepancies between the hardware implementation may go 
unnoticed from the quantization algorithm in the NN training framework (PyTorch in our case) which does not use TVM for the forward and backward propagation.
To avoid any such issue, we made sure that the results between TVM and PyTorch match for every single layer and stage to machine-precision accuracy, and we verified
the final Top-1 accuracy when executed in the hardware with integer-only arithmetic.
In~\appref{sec:error_accumulation_of_fq}, we present the error accumulation of feature maps for ResNet50 with INT4 quantization, which uses fake quantization in PyTorch and is deployed in TVM.

\section{Results}
\label{sec:results}

In this section, we first discuss ImageNet results on various models 
(ResNet18/50 and InceptionV3) for INT8, INT4, and mixed-precision INT4/8 with/without distillation.
Afterward, we study the different use cases of the ILP formulation, and
the corresponding trade-offs between model size, latency, and accuracy.
Detailed discussion on the implementation and set up is provided in~\appref{sec:exp_details}.
For all the experiments we made sure to report and compare with the highest
accuracy known for the baseline NN model in FP32 (i.e., we use a strong baseline for comparison). 
This is important since using a weak baseline accuracy could lead to misleading
quantization accuracy.

% ----------------------------------------------------------------
\begin{table}[!ht]
\caption{Quantization results for ResNet18/50 and InceptionV3.
Here, we abbreviate Integer-Only Quantization as ``Int'', Uniform Quantization as ``Uni'', the Baseline Accuracy as "BL", Weight Precision and Activation Precision as ``Precision'', Model Size as ``Size'' (in MB), Bit Operations as ``BOPS'' (in G), and Top-1 Accuracy as ``Top-1''. 
Also, ``WxAy'' means weight with x-bit and activation with y-bit, and 4/8 means mixed precision with 4 and 8 bits. ``MP'' means mixed precision with bitwidth ranging from 1-bit to 8-bit, and ``W1*'' means the bitwidth is 1-bit but the network architecture is changed (by using more channels). 
Our result with/without distillation is represented as \OURSD/\OURS.
}
\label{tab:imagenet}
\centering
\subfloat[\footnotesize ResNet18]{
\begin{adjustbox}{width=1\linewidth} 
\centering
\small
\setlength\tabcolsep{1.pt}
\begin{tabular}{lccccccccccccc} \toprule
Method                              &Int    & Uni       &  {BL}     & Precision     &Size       & BOPS      &Top-1\\
\midrule
\hc Baseline                        &\xm    & --        &   71.47   &   W32A32      &  44.6     &1858       &71.47 \\
\midrule    
\ha RVQuant~\cite{park2018value}    &\xm    &   \xm     &   69.91   &   W8A8        &  11.1     &116        & 70.01 \\
\hc \OURS                           &\cm    &   \cm     &   71.47   &   W8A8        &  11.1     &116        &\bf{71.56} \\
\midrule    
\ha PACT~\cite{choi2018pact}        &\xm    &   \cm     &   70.20   &   W5A5        &  7.2      &50         & 69.80 \\
\ha LQ-Nets~\cite{zhang_2018_lqnets}&\xm    &   \xm     &   70.30   &   W4A32       &  5.8      &225        & 70.00 \\
\hc \OURS                           &\cm    &   \cm     &   71.47   &   W4/8A4/8    &  6.7      &72         & 70.22\\
\hc \OURSD                          &\cm    &   \cm     &   71.47   &   W4/8A4/8    &  6.7      &72         & \bf{70.38}\\
\midrule
\ha CalibTIB\cite{hubara2020improving}&\xm  &   \cm     &   71.97   &   W4A4        &  5.8      &34         & 67.50\\
\hc \OURS                           &\cm    &   \cm     &   71.47   &   W4A4        &  5.8      &34         & \bf{68.45} \\
\bottomrule 
\end{tabular}
\end{adjustbox}
\label{tab:resnet18}
}

% \begin{subtable}
\subfloat[\footnotesize ResNet50]{
\begin{adjustbox}{width=1\linewidth} 
\centering
\small
\setlength\tabcolsep{1.pt}
\begin{tabular}{lccccccccccccc} \toprule
Method                              &Int            & Uni       &  {BL}         & Precision     &Size       & BOPS      &Top-1\\
\midrule
\hc Baseline                        &\cm            &  \cm      &   77.72       &   W32A32      &  97.8     &3951       &77.72 \\
\midrule
\ha Integer Only~\cite{jacob2018quantization} &\cm  &  \cm      &   76.40       &   W8A8        &  24.5     &247        &74.90      \\
\ha RVQuant~\cite{park2018value}    &\xm            &  \xm      &   75.92       &   W8A8        &  24.5     &247        &75.67\\
\hc \OURS                           &\cm            &  \cm      &   77.72       &   W8A8        &  24.5     &247        &\bf{77.58}\\
\midrule
\ha PACT~\cite{choi2018pact}        &\xm            &  \cm      &   76.90       &   W5A5        &  16.0     &101        &76.70 \\
\ha LQ-Nets~\cite{zhang_2018_lqnets}&\xm            &  \xm      &   76.50       &   W4A32       &  13.1     &486        &76.40 \\
\ha RVQuant~\cite{park2018value}    &\xm            &  \xm      &   75.92       &   W5A5        &  16.0     &101        &75.60 \\
\ha HAQ~\cite{wang2018haq}          &\xm            &  \xm      &   76.15       &   WMPA32      &  9.62     &520        &75.48 \\
\ha OneBitwidth~\cite{chin2020one}  &\xm            &  \cm      &   76.70       &   W1*A8       &  12.3     &494        &76.70 \\
\hc \OURS                           &\cm            &  \cm      &   77.72       &   W4/8A4/8    &  18.7     &154        & 75.39\\
\hc \OURSD                          &\cm            &  \cm      &   77.72       &   W4/8A4/8    &  18.7     &154        & \bf{76.73}\\
\midrule
\ha CalibTIB\cite{hubara2020improving} &\xm         &  \cm      &   77.20       &   W4A4        &  13.1     &67         & 73.70\\
\hc \OURS                           &\cm            &  \cm      &   77.72       &   W4A4        &  13.1     &67         & \bf{74.24}\\
\bottomrule 
\end{tabular}
\end{adjustbox}
\label{tab:resnet50}
}

\subfloat[\footnotesize InceptionV3]{
\begin{adjustbox}{width=1\linewidth} 
\centering
\small
\setlength\tabcolsep{1.pt}
\begin{tabular}{lccccccccccccc} \toprule
Method                              &Int            & Uni       &  {BL}         & Precision     &Size       & BOPS      &Top-1\\
\midrule
\hc Baseline                        &\xm            &   \cm     &   78.88       &   W32A32      &  90.9     &5850       &78.88 \\
\midrule
\ha Integer Only~\cite{jacob2018quantization}&\cm   &   \cm     &   78.30       &   W8A8        &  22.7     &366        &74.20 \\
\ha RVQuant~\cite{park2018value}    &\xm            &   \xm     &   74.19       &   W8A8        &  22.7     &366        & 74.22\\
\hc \OURS                           &\cm            &   \cm     &   78.88       &   W8A8        &  22.7     &366        &\bf{78.76}\\
\midrule
\ha Integer Only~\cite{jacob2018quantization}&\cm   &   \cm     &   78.30       &   W7A7        &  20.1     &280        &73.70 \\
\hc \OURS                           &\cm            &   \cm     &   78.88       &   W4/8A4/8    &  19.6     &265        & 74.65\\
\hc \OURSD                          &\cm            &   \cm     &   78.88       &   W4/8A4/8    &  19.6     &265        & \bf{74.72}\\
\midrule
\hc \OURS                           &\cm            &   \cm     &   78.88       &   W4A4        &  12.3     &92         &70.39\\
\bottomrule 
\end{tabular}
\label{tab:inceptionv3}
\end{adjustbox}
}
\end{table}
%%%%%%%%%%%%%%%%%%%%%%%

% ---------------------------------------------------------
\subsection{Low Precision Integer-Only Quantization Results}
\label{subsec:dyadic_results}

We first start with ResNet18/50 and InceptionV3 quantization on ImageNet, and compare the performance
of \OURS with other approaches, as shown in~\tref{tab:imagenet}.

\textbf{Uniform 8-bit Quantization.} 
Our 8-bit quantization achieves similar accuracy compared to
the baseline.
Importantly, for all the models \OURS achieves higher accuracy
than the integer-only approach of~\cite{jacob2018quantization}.
For instance, on ResNet50, we achieve  2.68\% higher accuracy as compared to~\cite{jacob2018quantization}. 
This is in part due to our BN folding strategy that was described in~\sref{subsec:batchnorm}.

\textbf{Uniform 4-bit Quantization.} 
To the best of our knowledge, 4-bit results of \OURS are the first integer-only quantization results
reported in the literature. 
The accuracy results for ResNet18/50, and InceptionV3 are quite high, despite the fact that all of the inference computations are restricted
to be integer multiplication, addition, and bit shifting.
While there is some accuracy drop,
this should not be incorrectly interpreted
that uniform INT4 is not useful. On the contrary, one has to keep in mind that
certain use cases have strict latency and memory footprint limit for which this may
be the best solution.
However, higher accuracy can be achieved through mixed-precision quantization.

\textbf{Mixed 4/8-bit Quantization.}
The mixed-precision results improve the accuracy by several percentages for all the
models, while slightly increasing the memory footprint of the model.
For instance, the mixed-precision result for ResNet18 is 1.88\% higher than its INT4 counterpart with just a 1.9MB increase in model size.
Further improvements are also possible with distillation (denoted as \OURSD in the table).
For ResNet50, the distillation can boost the mixed-precision by 1.34\%.
We found that distillation helps most for mixed-precision quantization, and we found little to no improvement
for uniform INT8, or uniform INT4 quantization cases.\footnote{We used simple distillation without
extensive tuning. One might be able to improve the results further with more sophisticated distillation algorithms.}

Overall, the results show that \OURS achieves comparable accuracy to prior quantization methods
including both uniform and mixed-precision quantization (e.g., PACT, RVQuant, OneBitwidth, HAQ which use FP32 arithmetic,
and/or non-standard bit precision such as 5 bits, or different bit-width for weights and activations).
Similar observations hold for InceptionV3, as reported in~\tref{tab:inceptionv3}.

%%%%%%%%%%%%%%%%%%%%%%%%%%%%%%%%%%%%%%%%%%%%%%%%%%%%%
\begin{table}[!ht]
\caption{Mixed-precision quantization results for ResNet18 and ResNet50 with different constraints. 
Here, we abbreviate constraint level as ``Level''.
Model Size as ``Size'', Bit Operations as ``BOPS'', 
the speedup as compared to INT8 results as ``Speed'', and Top-1 Accuracy as ``Top-1'',
The last column of Top-1 represents results of \OURS and \OURSD. 
Note that for uniform INT8 ResNet50 (ResNet18), the latency is 1.06ms (0.40ms) per images.
}
\label{tab:resnet18-50-constraint}
\centering
\subfloat[\footnotesize ResNet18]{
\centering
\small
\setlength\tabcolsep{3.pt}
\begin{tabular}{c|cccccccccccccc}
\toprule
                    &Level  & Size (MB)  & BOPS (G) & Speed &Top-1 \\
\midrule
INT8                &--& 11.2       & 114      & 1x     &71.56\\
\midrule
\parbox[t]{2mm}{\multirow{3}{*}{\rotatebox[origin=c]{90}{Size}}} 
                    & High       &\bf{ 9.9}             & 103        & 1.03x       & 71.20/71.59\\
                    &\chb Medium &\chb \bf{7.9}        &\chb 98    &\chb 1.06x   &\chb 70.50/71.09\\
                    &\chc Low    &\chc \bf{7.3}        &\chc 95    &\chc 1.08x   &\chc 70.01/70.66\\
\midrule
\parbox[t]{2mm}{\multirow{3}{*}{\rotatebox[origin=c]{90}{BOPS}}} 
                    & High   & 8.7           & \bf{92}        & 1.12x        & 70.40/71.05\\
                    &\chb Medium &\chb 6.7       &\chb \bf{72}    &\chb 1.21x    &\chb 70.22/70.38\\
                    &\chc Low    &\chc 6.1       &\chc \bf{54}    &\chc 1.35x    &\chc 68.72/69.72\\
\midrule
\parbox[t]{2mm}{\multirow{3}{*}{\rotatebox[origin=c]{90}{Latency}}} 
                    & High   & 8.7           & 92        & \bf{1.12x}       &70.40/71.05\\ % the same as bops 0.75
                    &\chb Medium &\chb 7.2       &\chb 76    &\chb \bf{1.19x}   &\chb 70.34/70.55\\
                    &\chc Low    &\chc 6.1        &\chc 54    &\chc \bf{1.35x}   &\chc 68.56/69.72\\ % the same as bops 0.25
\midrule 
INT4                & --& 5.6        & 28    & 1.48x   & 68.45\\
\bottomrule
\end{tabular}
\label{tab:constraint_resnet18}
}

\subfloat[\footnotesize ResNet50]{
\centering
\small
\setlength\tabcolsep{3.pt}
\begin{tabular}{l|cccccccccccccc}
\toprule
                    &Level & Size (MB)  & BOPS (G) & Speed &Top-1 \\
\midrule 
INT8                &--& 24.5       & 247      & 1x     &77.58\\
\midrule
\parbox[t]{2mm}{\multirow{3}{*}{\rotatebox[origin=c]{90}{Size}}} 
                    & High       & \bf{21.3}           & 226        & 1.09x   & 77.38/ 77.58\\
                    &\chb Medium &\chb \bf{19.0}       &\chb 197    &\chb 1.13x  &\chb 75.95/76.96 \\
                    &\chc Low    &\chc \bf{16.0}       &\chc 168    &\chc 1.18x  &\chc 74.89/76.51\\
\midrule
\parbox[t]{2mm}{\multirow{3}{*}{\rotatebox[origin=c]{90}{BOPS}}} 
                    & High       & 22.0           & \bf{197}        & 1.16x       & 76.10/76.76 \\
                    &\chb Medium &\chb 18.7       &\chb \bf{154}    &\chb 1.23x  &\chb 75.39/76.73\\
                    &\chc Low    &\chc 16.7       &\chc \bf{110}    &\chc 1.30x  &\chc 74.45/76.03\\
\midrule
\parbox[t]{2mm}{\multirow{3}{*}{\rotatebox[origin=c]{90}{Latency}}} 
                    & High       & 22.3           & 199        & \bf{1.13x}      & 76.63/76.97\\
                    &\chb Medium &\chb 18.5       &\chb 155    &\chb \bf{1.21x}  &\chb 74.95/76.39 \\
                    &\chc Low    &\chc 16.5       &\chc 114    &\chc \bf{1.28x}  &\chc 74.26/76.19\\
\midrule 
INT4                & --& 13.1        & 67    & 1.45x   & 74.24\\
\bottomrule
\end{tabular}
\label{tab:constraint_resnet50}
 }
\end{table}
%%%%%%%%%%%%%%%%%%%%%%%%%%%%%%%%%%%%%%%%%%%%%%%%%%%%%

% ----------------------------------------------------------------------------------
\subsection{Mixed-precision Results with Different Constraints}
\label{sec:mixed_precision_with_diff_constraints}

Here, we discuss various scenarios where different constraints could
be imposed for quantization, and the interesting trade-offs associated with each scenario.
The ILP problem in~\eref{eq:ilp} has three constraints of 
model size, BOPS, and latency. 
We consider three different thresholds
for each of the constraints and study how the ILP balances the trade-offs to obtain an optimal
quantized model.
We also focus on the case, where the practitioner is not satisfied with the
performance of the INT4 quantization and wants to improve the performance (accuracy, speed, and 
model size) through mixed-precision quantization (INT4 and INT8).
The ILP formulation enables the practitioner to set each or all of these constraints.
Here, we present results when only one of these constraints is set at a time.
The results are shown in~\tref{tab:resnet18-50-constraint}, which is split into three sections of Size (model size), BOPS, and Latency. Each section represents the corresponding
constraint as specified by the practitioner. The ILP solver then finds the optimal
mixed-precision setting to satisfy that constraint, while maximizing accuracy. 
See~\appref{sec:ilp_interpolation} for the example of the latency constraint for ResNet18.

We start with the model size and BOPS constraints for ResNet18.
The model size of pure INT4 quantization is 5.6MB, and INT8 is 11.2MB.
However, the accuracy of INT4 quantization is 68.45\% which maybe low
for a particular application. The practitioner then has the option to
set the model size constraint to be slightly higher than pure INT4.
One option is to choose 7.9MB which is almost in between INT4 and INT8.
For this case, the ILP solver finds a bit-precision setting that results
in 71.09\% accuracy which is almost the same as INT8. This model
is also 6\% faster than INT8 quantization.

Another possibility is to set the speed/latency as a constraint.
The results for this setting are represented under the ``Latency'' row in~\tref{tab:resnet18-50-constraint}.
For example, the practitioner could request the ILP to find a bit-precision
setting that would result in 19\% faster latency as compared to the INT8 model (see ``Medium'' row).
This results in a model with an accuracy of 70.55\% with a model size of only 7.2MB.
A similar constraint could also be made for BOPS.

Several very interesting observations can be made from these results.
(i) The correlation between model size and BOPS is weak which is expected. That is a larger model size does not mean higher BOPS and vice versa.
For example, compare Medium-Size and High-BOPS for ResNet18.
The latter has lower BOPS despite being larger (and is actually faster as well).
(ii) The model size does not directly correlate with accuracy. 
For example, for ResNet50, High-BOPS has a model size of 22MB and accuracy
of 76.76\%, while High-Size has a smaller model size of 21.3MB but higher accuracy of 77.58\%.

In summary, although directly using INT4 quantization may result in large accuracy degradation, we can achieve significantly improved
accuracy with much faster inference as compared to INT8 results. This gives the practitioner a wider range of choices beyond
just INT8 quantization.
Finally, we should mention that the accuracy and speed for all of the results shown for ResNet18/50 and InceptionV3 have been verified
by directly measuring them when executed in quantized precision in hardware through TVM. As such, these results are actually
what the practitioner will observe, and these are not simulated results.

\section{Conclusions}
\label{sec:conclusions}
In this work, we presented \OURS, a new low-precision integer-only quantization framework, 
where the entire inference is executed with only integer multiplication, addition,
and bit shifts. In particular, no FP32 arithmetic or even integer division is used in the entire inference.
We presented results for uniform and mixed-precision INT4/8.
For the latter, we proposed a hardware-aware ILP based method that
finds the optimal trade-off between model perturbation and application 
specific constraints such as model size, inference speed, and total BOPS.
The ILP problem can be solved very efficiently, under a second for all the models considered here.
We showed that our approach can achieve up to 5\% higher accuracy as
compared to the prior integer-only approach of~\cite{jacob2018quantization}.
Finally, we directly implemented the low-precision quantized models
in hardware by extending TVM to support INT4 and INT4/8 inference.
We verified all the results, by matching the activation of each layer
with our PyTorch framework (up to machine precision), including the verification of
the final accuracy of the model.
The framework, the TVM implementation, and the quantized models have been open sourced~\cite{HAWQ}.

\section*{Acknowledgments}
The UC Berkeley team also acknowledges gracious support from Samsung (in particular Joseph Hassoun), Intel corporation, Intel VLAB team, Google TRC team, and Google Brain (in particular Prof. David Patterson, Dr. Ed Chi, and Jing Li).
Amir Gholami was supported through through funding from Samsung SAIT.
Michael W. Mahoney would also like to acknowledge the UC Berkeley CLTC, ARO, NSF, and ONR.
Our conclusions do not necessarily reflect the position or the policy of our sponsors, and no official endorsement should be inferred.

% \clearpage
% In the unusual situation where you want a paper to appear in the
% references without citing it in the main text, use \nocite
%\nocite{langley00}
{
\bibliography{ref}
\bibliographystyle{icml2021}
}

\clearpage
\appendix

%%%%%%%%%%%%%%%%%%%
% Re-count the Figure/Algorithm/Tables after this point. 
%%%%%%%%%%%%%%%%%%%
\counterwithin{figure}{section}
\counterwithin{table}{section}

% --------------------------------------------
\section{Deployment Frameworks}
\label{sec:deplotment_frameworks}
A number of frameworks~\cite{jia2014caffe, chen2015mxnet,abadi2016tensorflow,seide2016cntk, paszke2017automatic, gulli2017deep, vasilache2018tensor, chen2018tvm} have 
been developed for deep learning. Many~\cite{jia2014caffe, chen2015mxnet, abadi2016tensorflow, paszke2017automatic} offer a dataflow DAG abstraction for specifying NN workloads and provide optimization support for inference as well as training with automatic differentiation. 
These frameworks significantly reduce development cycles for deep learning algorithms and thus facilitate innovations in deep learning. 
However, a majority of these frameworks~\cite{jia2014caffe, chen2015mxnet, paszke2017automatic} adopt a library-based approach that maps the NN operations to 
hardware through existing high-performance libraries, such as cuDNN~\cite{chetlur2014cudnn} for GPUs, and GEMMLOWP~\cite{jacob2017gemmlowp} and NNPACK~\cite{dukhan2016nnpack} for CPUs. 
These libraries currently do not support low-precision inference (INT4), 
and since they are not open source we could not add that functionality.
As such, for our analysis we adopted to use TVM~\cite{chen2018tvm},
which provides 
a general graph and a tensor expression intermediate representation (IR)
to support automatic code transformation and generation. 
TVM also equips a QNN dialect~\cite{jain2020efficient} to compile the quantization-specific operators of a quantized model.
We choose TVM as our deployment framework for several reasons including:
(i) its extensive support in the frontend high-level frameworks and the backend hardware platforms; and 
(ii) its decoupled IR abstraction that separates the algorithm specifications and the 
scheduling decisions.
Augmenting TVM with our mixed-precision quantization support allows this optimization 
to be used by NNs written in different frameworks as well as for various target 
hardware platforms.   
In addition, the decoupled IR design in TVM allows the mixed-precision quantization 
optimization to be applied without affecting the specification of algorithms.

\section{Quantization Method}
\label{sec:quantization_method}

\textbf{Symmetric and Asymmetric Quantization.} 
For uniform quantization, the scaling factor $S$ is chosen to equally partition the range of real
values $r$ for a given bit width:
\begin{equation*}
\small
    S = \frac{r_{max} - r_{min}}{2^{b} - 1},
\end{equation*}
where $r_{max},\ r_{min}$ denotes the max/min value of the real values, and $b$ is the quantization bit
width. This approach is referred to as \emph{asymmetric quantization}.
It is also possible to use a \emph{symmetric quantization} scheme
where $S = {2\max(|r_{max}|, |r_{min}|)}/{(2^b-1)}$ and $Z=0$ (since zero will be exactly represented).
As such, the quantization mapping can be simplified as:
\begin{equation}
\small
\label{eq:quantization_formula_our}
Q(r) = \text{Int}\left(\frac{r}{S}\right).
\end{equation}
Conversely, the real values $r$ could be recovered from the quantized values $Q(r)$ as follows:
\begin{equation}
\small
    \tilde r = S~Q(r).
\end{equation}
Note that the recovered real values $\tilde{r}$ will not exactly
match $r$ due to the rounding operation. 
For \OURS, we use symmetric quantization for weights and asymmetric quantization for the activations.

\textbf{Static and Dynamic Quantization.}
The scaling factor $S$ depends on $r_{max}$ and $r_{min}$.
These can be precomputed for weights. However, for activations, each input will have a 
different range of values across the NN layers.
In dynamic quantization, this range and the corresponding scaling factor is computed for each activation map during runtime. 
However, computing these values during inference has high
overhead. 
This can be addressed with static quantization, in which this range is pre-calculated during the quantization phase and made independent of the input data, by analyzing the range of activations for different batches.
We use static quantization for all of the experiments with \OURS.
With these definitions, we next discuss how quantized inference is performed.

% ------------------------------------------------------------
\section{Fake Quantization for Convolution}
\label{sec:fake_quantization}

\begin{figure*}[t]
\centering
\includegraphics[width=.49\textwidth]{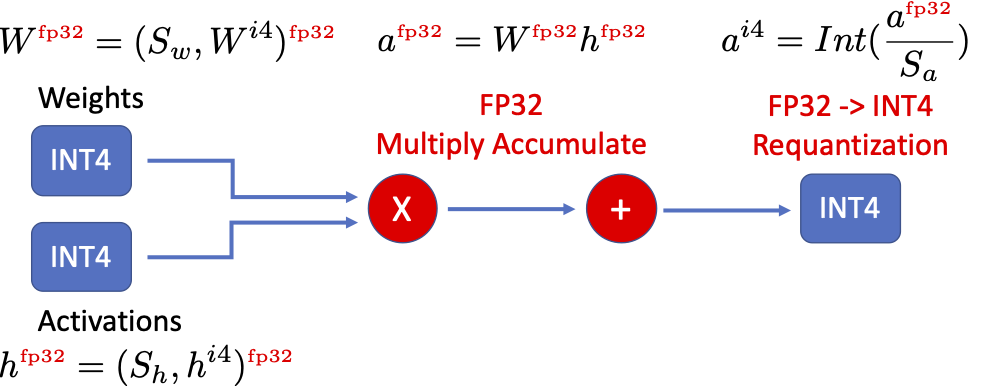}
\includegraphics[width=.49\textwidth]{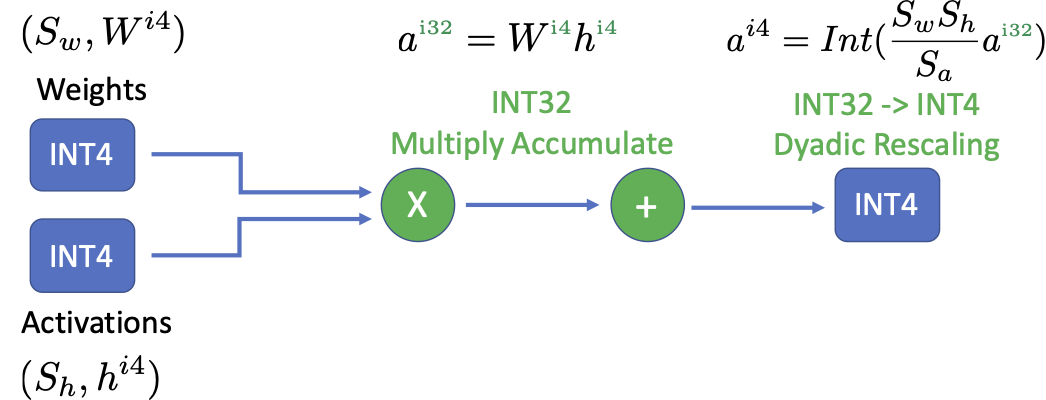}
\caption{
Illustration of fake vs true quantization for a convolution (fully-connected) layer. 
(Left) In the simulated quantization (aka fake quantization), weights and activations
are simulated as integers with floating point representation, and all the multiplication 
and accumulation happens
in FP32 precision.
However, with this approach, one cannot benefit from low-precision ALUs.
(Right) An illustration of the integer-only pipeline with integer-only quantization.
Note that with this approach, all the weights and activations
are stored in integer format, and all the multiplications are performed with INT4 and accumulated in INT32 precision. 
Finally, the accumulated result is requantized to INT4 with dyadic scaling (denoted by $(\frac{S_wS_h}{ S_a}$)). Importantly, no floating point or even
integer division is performed.
}
\label{fig:int32_accumulation}
\end{figure*}

In simulated quantization (also referred to as fake quantization in literature), all the calculations happen in FP32, which is different from the approach we used in~\sref{sec:quantized_matmul}. 
Similar to \sref{sec:quantized_matmul}, suppose that the hidden activation is $h=S_hq_h$ and weight tensor is $W=S_wq_w$.
In fake quantization, the output is calculated as:
\begin{equation}
\small
    a = (S_wq_w) * (S_hq_h).
\end{equation}
That is the weight and activation are first represented back to FP32 precision,
and then the calculation is performed. 
This result is then requantized and sent to the next layer as follows:
\begin{equation}
% \label{eq:requantize_full}
q_a =\text{Int}\left(\frac{a}{S_a}\right),
\end{equation}
where $S_a$ is the pre-calculated scale factor for the output activation.
However, notice that here the requantization operation requires FP32 arithmetic (division by $S_a$), which is different from \OURS's Dyadic arithmetic that only uses integer operations. 
Figure~\ref{fig:int32_accumulation} shows the illustration of fake vs true quantization for a convolution (fully-connected) layer, without the BN layer. We also
showed the corresponding illustration when BN is used in~\fref{fig:int32_batchnorm}.

%%%%%%%%%%%%%%%%%%%%%%%%%%%%%%%%%%%%%%%%%%%%%%%%%%%%%%%%%%%%%%%%%%%%%%%%%%%%%%%%%%%
%%%%%%%%%%%%%%%%%%%%%%%%%%%%%%%%%%%%%%%%%%%%%%%%%%%%%%%%%%%%%%%%%%%%%%%%%%%%%%%%%%%

% -----------------------------------------------------------
\section{Batch Normalization Fusion}
\label{sec:batchnorm_fusion}

During inference, the mean and standard deviation used in the BN layer are the running statistics (denoted as $\mu$ and $\sigma$). 
Therefore, the BN operation can be fused into the previous convolutional layer. 
That is to say, we can combine BN and CONV into one operator as,
\begin{equation}
\small
\label{eq:conv_bn}
\begin{split}
    \text{CONV\_BN}(h) 
    &= \beta \frac{Wh - \mu}{\sigma} + \gamma \\
    &= \frac{\beta W}{\sigma}h + (\gamma - \frac{\beta\mu}{\sigma}) \equiv \bar W h + \bar b, 
\end{split}
\end{equation}
where $W$ is the weight parameter of the convolution layer and $h$ is the input feature map. 
In \OURS, we use the fused BN and CONV layer and quantize $\bar W$ to 4-bit or 8-bit based on the setting, and quantize the bias term, $\bar b$ to 32-bit. 
More importantly, suppose the scaling factor of $h$ is $S_h$ and the scaling factor of $\bar W$ is $S_{\bar W}$. 
The scaling factor of $\bar b$ is enforced to be 
\begin{equation}
\small
S_{\bar b} = S_h S_{\bar W}.
\end{equation}
So that the integer components of $\bar W h$ and $\bar b$ can be directly added during inference. 

% -------------------------------------------
\section{Concatenation Layer}
\label{sec:concatenation_layer}
The concatenation operation in Inception is an important component, which needs to be quantized carefully to avoid significant accuracy degradation.
Concatenation layers are often used in the presence of pooling layers and other convolutions
(a good example is the inception family of NNs).
In \OURS, we use INT32 for the pooling layer since performing pooling on 4-bit can result in significant information loss. Furthermore, we perform separate dyadic arithmetic for the following concatenation operator in the inception module. 
Suppose the input of a concatenation block is denoted as $h=S_{h}q_{h}$, the output of the three 
convolutional branches are $m=S_{m}q_{m}$, $n=S_{n}q_{n}$, and $l=S_lq_{l}$, the output of the pooling branch is $p=S_{p}q_{p}$, and the 
final output is $a=S_{a}q_{a}$. 

\begin{figure}[t]
\centering
\includegraphics[width=0.46\textwidth]{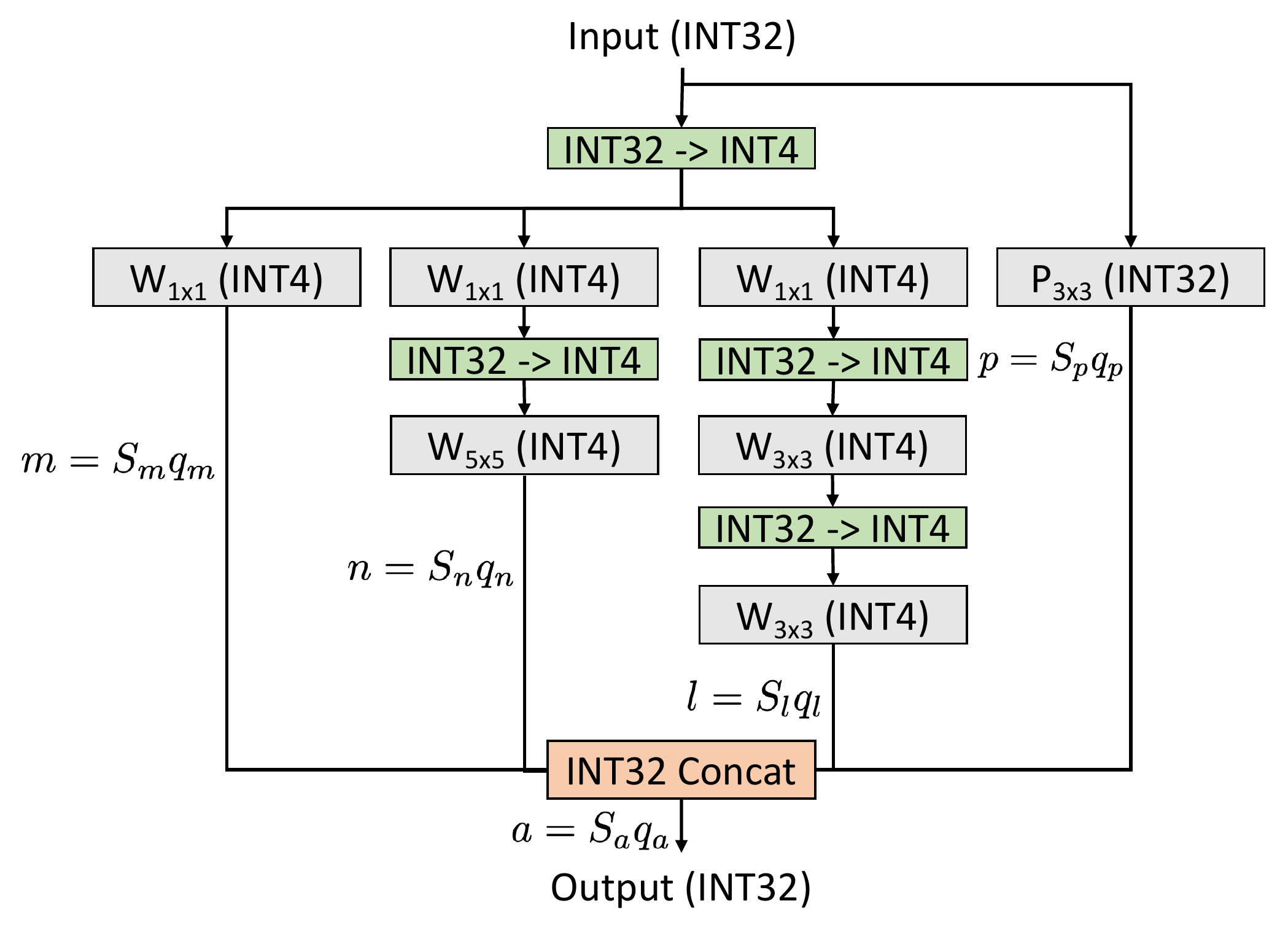}
\caption{
Illustration of \OURS for an inception module.
Input feature map is given in INT32 precision, which is requantized to
INT4 precision (green boxes) before 
being passed to the three convolutional branches. 
The pooling layer, however, is performed on the original input feature
map in INT32. This is important since performing pooling on 4-bit data
can result in significant information loss.
The outputs for all the branches are scaled and requantized before being
concatenated.
}
\label{fig:integer_only_inception}
\end{figure}

The pooling branch directly takes $h$ as input, and the rest of the three convolutional branches take the quantized 4-bit tensor as input. 
After the computation of four separate branches, the output $q_{a}$ is calculated with four DN operators:
\begin{equation}
\small
    q_{a} = \sum_{i\in\{m,n,l\}}\text{DN}\left(\frac{S_{i}}{S_{a}}\right) q_i + \text{DN}\left(\frac{S_{p}}{S_{a}}\right) q_p.
\end{equation}
This scheme is represented in~\fref{fig:integer_only_inception}.

%%%%%%%%%%%%%%%%%%%%%%%%%%%%%%%%%%%%%%%%%%%%%%%%%%%%%%%%%%%%%%%%%%%%%%%%%%%%%%%%%
%%%%%%%%%%%%%%%%%%%%%%%%%%%%%%%%%%%%%%%%%%%%%%%%%%%%%%%%%%%%%%%%%%%%%%%%%%%%%%%%%

% -----------------------------------------------------------------
\section{Fake Quantization for Residual Connection}
\label{sec:fake_res_con_layer}

Similar to~\sref{sec:residual_connection}, 
Let us denote the activation passing through the residual connection as
$r=S_{r}q_{r}$.
the activation of the main branch before residual addition as
$m=S_{m}q_{m}$. the final output after
residual accumulation as $a=S_{a}q_{a}$. 
In fake quantization, the output $a$ is calculated in FP32 as,
\begin{equation}
\small
    a = S_rq_r + S_mq_m.
\end{equation}
Afterwards, requantization is performed,
\begin{equation}
\small
    q_a = \text{Int}(\frac{S_rq_r + S_mq_m}{S_a}),
\end{equation}
where the $\text{Int}$ operator requires FP32 multiplication. 

Similarly, fake quantization for concatenation layer is calculated as (see~\appref{sec:concatenation_layer} for notations):
\begin{equation}
\small
    q_{a} = \text{Int}(\frac{m+n+l+p}{S_a}).
\end{equation}

% ----------------------------------------------------------
\section{Error Accumulation of Fake Quantization}
\label{sec:error_accumulation_of_fq}

%%%%%%%%%%%%%%%%%%%%%%%%%%%%%%%%%%%%%%%%%%%%%%%%%%%%%%%%%%%%%%%%%%%%%%%%%%%%%%%%%
\begin{figure}[t]
\centering
\includegraphics[trim=30 25 20 20, clip, width=0.52\textwidth]{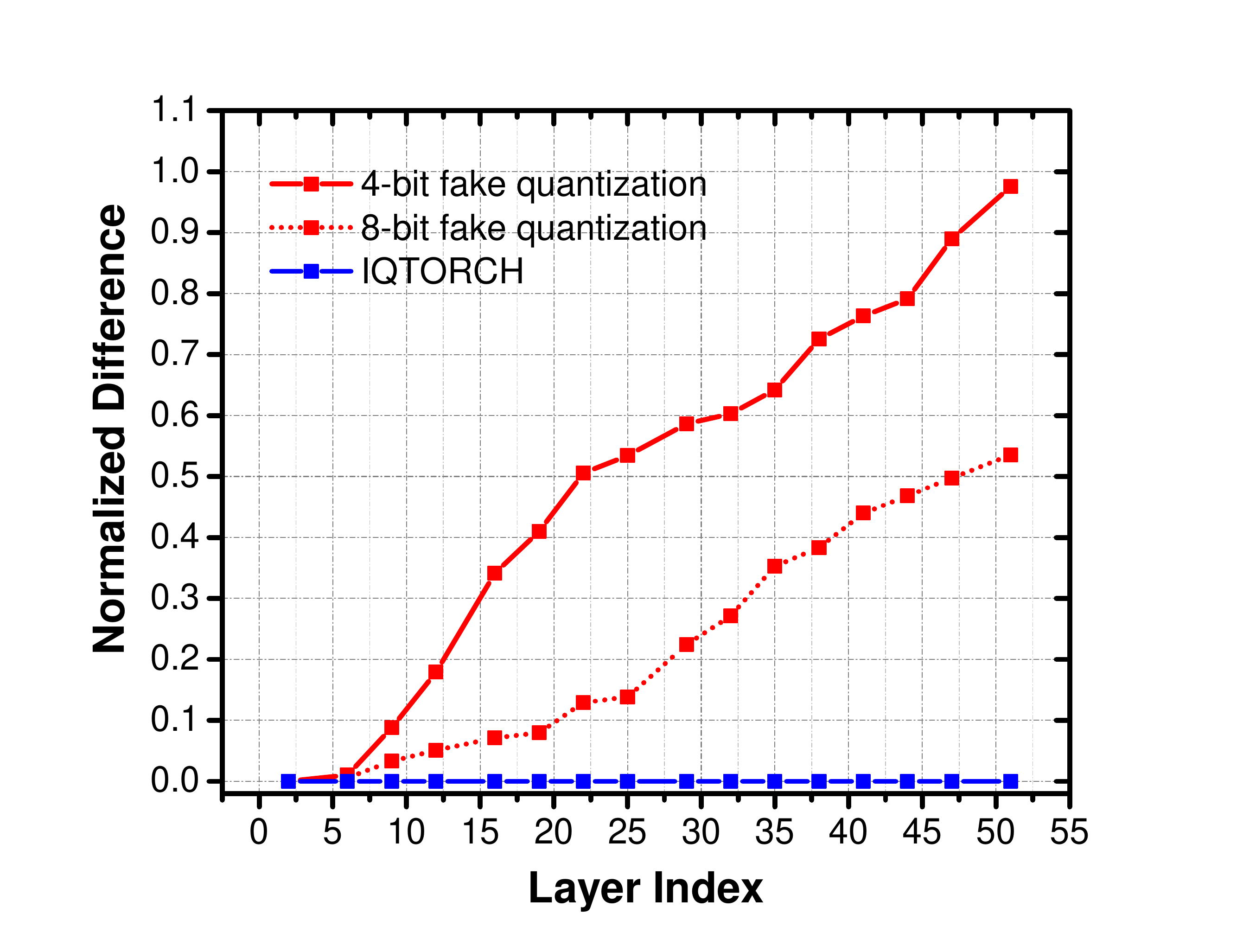}
\caption{The normalized difference between activation tensors in TVM and activation tensors in 
PyTorch during inference. The normalized difference is the $L_2$ norm of the difference between two activation counterparts divided by the $L_2$ norm of the TVM activation tensor.}
\label{fig:accumulation}
\end{figure}
%%%%%%%%%%%%%%%%%%%%%%%%%%%%%%%%%%%%%%%%%%%%%%%%%%%%%%%%%%%%%%%%%%%%%%%%%%%%%%%%%

There has been a common misunderstanding that
using fake quantization is acceptable since one
can use FP32 precision to perform Integer operations exactly.
First, this is only true if the matrix multiplications
only use integer numbers, without using very large numbers.
The latter is the case in most ML applications.
However, the problem is that many quantization approaches use
fake quantization in a way that is different than the above argument.

For example, keeping the BN parameters in FP32 and not quantizing them
is a major problem. It is not possible to simply ignore that and deploy
a quantized model with FP32 BN parameters on integer-only hardware.
This difference was discussed and illustrated in~\fref{fig:int32_batchnorm}.

Another very important subtle issue is how the residual connection is treated.
As discussed in the previous section, the fake quantization approaches use
FP32 arithmetic to perform the residual addition. The common (but incorrect) 
argument here again is that the INT arithmetic can be performed without error with
FP32 logic. However, this is not the problem, since there is a subtle difference
in how requantization is performed.
In fake quantization, the results are first accumulated in FP32 and then requantized.
However, it is not possible to perform such an operation on integer-only hardware, where the results are always quantized and then accumulated.
This difference can actually lead to O(1) error.

For example consider the following case: assume $S_a=1$, $r=2.4$, $m=4.4$ (see definition in~\appref{sec:fake_res_con_layer}), and the requantization operator ($\text{Int}$) uses the ``round to the nearest integer''.
Then using fake quantization, the output $q_a$ is 
\begin{equation}
    \small 
    q_a = \text{Int}(4.4 + 2.4) =7.
\end{equation}
However for true quantization, the output $q_a$ is 
\begin{equation}
    \small 
    q_a = \text{Int}(4.4) + \text{Int}(2.4) = 6.
\end{equation}
This is an O(1) error that will propagate throughout the network. Also
note that the problem will be much worse for low precision error. This is because
an O(1) error for INT8 quantization is equivalent to a constant times (1/256), while
for INT4 quantization it will be a constant times (1/16).

We also performed a realistic example on ResNet50 for the uniform quantization
case. 
We perform fake quantization in PyTorch for fine-tuning and then deploy the model in TVM
using integer-only arithmetic.
Afterwards, we calculate the error between the feature map of PyTorch (fake quantization) and TVM (integer-only).
In particular, we measure the normalized difference using $L_2$ norm:
\begin{equation}
\small
    \text{Normalized\_Difference} = \frac{\|x_1 - x_2\|}{\|x_1\|},
\end{equation}
where $x_1,\ x_2$ are the feature maps with fake quantization and the corresponding values calculated
in hardware with integer-only arithmetic.
In~\fref{fig:accumulation} we show the normalized difference between activation tensors in TVM and activation tensors in PyTorch during inference. 
As one can see, the numerical differences of the first layers are relatively small. 
However, this error accumulates throughout the layers and becomes quite significant in the last layers. 
Particularly, for uniform 4-bit quantization, the final difference becomes $>95\%$.

% ---------------------------------------------------------
\section{Implementation Details}
\label{sec:exp_details}
\paragraph{Models}
All the empirical results are performed using pretrained models from
PyTorchCV~\cite{pytorchcv} library.
In particular, we do not make any architectural changes to the models, even though doing so might
lead to better accuracy.
We consider three NN models, ResNet18, ResNet50, and InceptionV3, trained on
the ImageNet dataset~\cite{deng2009imagenet}.
For all the NNs, we perform BN folding to speed up
the inference. 
All the calculations during inference are performed using dyadic arithmetic (i.e., integer addition, multiplication, and bit shifting),
with no floating point or integer division anywhere in the network, including
requantization stages.

\paragraph{Training details}
We use PyTorch (version 1.6) for quantizing models with \OURS.
For all the quantization results, we follow the standard practice of keeping the first and last layer in 8-bit (note that input data
is encoded with 8-bits for the RGB channels, which is quantized with symmetric quantization).
We only use uniform quantization along with channel-wise symmetric quantization for weights, and we use layer-wise asymmetric quantization for activations.
In order to perform static quantization, we set our momentum factor of quantization range (i.e., minimum and maximum) of activations to be 0.99 during training.
Although further hyperparameter tuning may achieve better accuracy, for uniformity, all our experiments are conducted using
learning rate 1e-4, weight decay 1e-4, and batch size~128.

\paragraph{Distillation}
As pointed out previously~\cite{polino2018model}, for extra-low bit quantization
(in our case uniform 4 bit and mixed 4/8 bit quantization), distillation may alleviate the performance degradation from quantization. 
Therefore, in addition to our basic results, we also present results with distillation (denoted with \OURSD).
Among other things, we do confirm the findings of previous work~\cite{polino2018model} that distillation can boost the accuracy of quantized models. 
For all different models, we apply ResNet101~\cite{he2016deep} as the teacher,
and the quantized model as the student. 
For simplicity, we directly use the naive distillation method proposed in~\cite{hinton2015distilling}.
(More aggressive distillation or fine-tuning with hyperparameter may lead to better results).

\paragraph{Latency Measurement}
We use TVM to deploy and tune the latency of
the quantized models using Google Cloud Platform virtual machines with Tesla T4 GPUs and CUDA 10.2.
We build the same NN models in TVM and tune the layerwise performance by using the autotuner.
Once we have the tuned models, we run the end-to-end inference multiple times to measure the average latency.
For the accuracy test, we load the parameters trained from PyTorch and preprocess it to the corresponding data layout that TVM requires.
Then, we do inference in TVM and verify that the final accuracy matches the results in~PyTorch.% 

\paragraph{Mixed-precision configuration}
For mixed-precision configuration, 
we first compute the trace of each layer~\cite{dong2019hawqv2} using PyHessian~\cite{yao2019pyhessian}, and then solve the ILP problem using PULP~\cite{pulp}.
Our mixed-precision ILP problem can find the right bit-precision configuration with orders of magnitude faster run time, as compared to the RL based method~\cite{wang2018haq,wu2018mixed}. 
For instance, the entire trace computation can be finished within 30 minutes for all layers of ResNet50/InceptionV3 with only 4 RTX 6000 GPUs.
Afterward, the ILP problem can be solved in \textbf{less than a second}
(on a 15 inch MacBook Pro), as compared to more than 10/50 hours searching using RL~\cite{wang2018haq} with 4 RTX 6000 GPUs.

% -------------------------------------------
\section{ILP Result Interpolation}
\label{sec:ilp_interpolation}

%%%%%%%%%%%%%%%%%%%%%%%%%%%%%%%%%%%%%%%%%%%%%%%%%%%%%
\begin{figure*}[ht]
    \centering
    \includegraphics[width=0.99\textwidth]{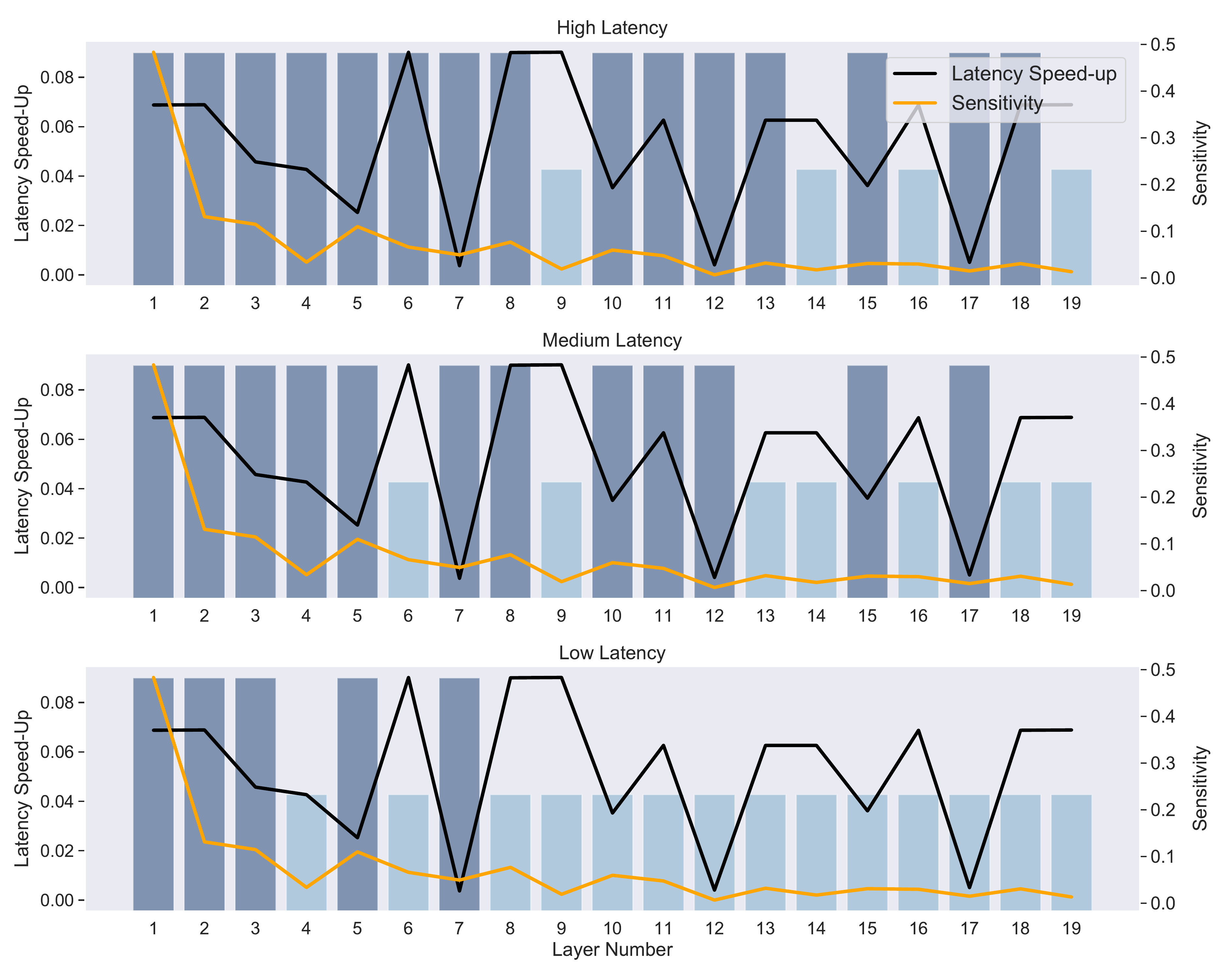}
    \caption{
    Illustration of the final model specification that the ILP solver finds
    for ResNet18 with latency constraint.
    The black line shows the percentage of latency reduction for
    a layer executed in INT4 versus INT8, normalized by total inference reduction.
    Higher values mean higher speedup with INT4.
    The orange line shows the sensitivity difference between INT8 and INT4 quantization using second order Hessian sensitivity~\cite{dong2019hawqv2}.
    The bit-precision setting found by ILP is shown in bar plots, with
    the blue and taller bars denoting INT8, and cyan and shorter 
    bars denoting INT4.
    Each row corresponds to the three results presented in~\tref{tab:constraint_resnet18}
    with latency constraint.
    For the low latency constraint, the ILP solver
    favors assigning INT4 for layers that exhibit large gains in latency when executed
    in INT4 (i.e., higher values in dark plot) and that have low sensitivity (lower values
    in the orange plot).
    }
    \label{fig:resnet18_conv_speedups}
\end{figure*}
%%%%%%%%%%%%%%%%%%%%%%%%%%%%%%%%%%%%%%%%%%%%%%%%%%%%%

We plot the bit-precision setting for each layer of ResNet18 that the ILP solver finds for different latency constraints, as shown in~\fref{fig:resnet18_conv_speedups}.
Additionally, we also plot the sensitivity ($\Omega_i$ in~\eref{eq:ilp}) and the corresponding speed up for each layer computed by quantizing the 
respective layer in INT8 quantization versus INT4.
As can be seen, the bit configuration chosen by the ILP solver is highly intuitive based
on the latency speed-up and the sensitivity. 
Particularly, when the mixed-precision model is constrained by the High-Latency setting (the first row of~\fref{fig:resnet18_conv_speedups}), 
only relatively insensitive layers, along with those that enjoy high INT4 speed-up, are quantized (i.e., layers 9, 14, and 19).
However, for the more strict Low-Latency setting (last row of~\fref{fig:resnet18_conv_speedups}),
only very sensitive layers are kept at INT8 precision (layer 1, 2, 3, 5, and~7).\footnote{Note that here layer 7 is the downsampling layer along with layer 5, so it is in the same bit setting as layer 5 even though the latency gain of layer 7 is limited.}

\end{document}